\def\paperlanguage{}
\pdfoutput=1 

\documentclass[letterpaper, 10 pt, conference]{ieeeconf}  

\usepackage{bm}
\usepackage{cite}
\usepackage{flushend}
\include{preamble}

\usepackage{booktabs}
\usepackage{siunitx}

\IEEEoverridecommandlockouts                              

\title{\LARGE \textbf
  {
    \switchlanguage%
    {%
    Vlimb: A Wire-Driven Wearable Robot for Bodily Extension, Balancing Powerfulness and Reachability
    }%
    {%
    力強さと動作範囲を両立する\\ワイヤ駆動型身体拡張ウェアラブルロボットVlimbの設計法
    }%
  }
}

\author{Shogo Sawaguchi$^{1}$, Temma Suzuki$^{1}$, Akihiro Miki$^{1}$, Kento Kawaharazuka$^{1}$,\\Sota Yuzaki$^{1}$, Shunnosuke Yoshimura$^{1}$, Yoshimoto Ribayashi$^{1}$, Kei Okada$^{1}$, Masayuki Inaba$^{1}$
  \thanks{$^{1}$ The authors are with the Department of Mechano-Informatics, Graduate School of Information Science and Technology, The University of Tokyo, 7-3-1 Hongo, Bunkyo-ku, Tokyo, 113-8656, Japan.
    {\texttt\small [sawaguchi, t-suzuki, miki, kawaharazuka, yuzaki, yoshimura, ribayashi, k-okada, inaba]@jsk.t.u-tokyo.ac.jp}
  }
}
\begin{document}

\maketitle
\thispagestyle{empty}
\pagestyle{empty}

\begin{abstract}
  \switchlanguage%
  {%
  Numerous wearable robots have been developed to meet the demands of physical assistance and entertainment.
  These wearable robots range from body-enhancing types that assist human arms and legs to body-extending types that have extra arms.
  This study focuses specifically on wearable robots of the latter category, aimed at bodily extension.
  However, they have not yet achieved the level of powerfulness and reachability equivalent to that of human limbs, limiting their application to entertainment and manipulation tasks involving lightweight objects.
  Therefore, in this study, we develop an body-extending wearable robot, Vlimb, which has enough powerfulness to lift a human and can perform manipulation.
  Leveraging the advantages of tendon-driven mechanisms, Vlimb incorporates a wire routing mechanism capable of accommodating both delicate manipulations and robust lifting tasks. 
  Moreover, by introducing a passive ring structure to overcome the limited reachability inherent in tendon-driven mechanisms, Vlimb achieves both the powerfulness and reachability comparable to that of humans.
  This paper outlines the design methodology of Vlimb, conducts preliminary manipulation and lifting tasks, and verifies its effectiveness.
  }%
  {%
  身体補助やエンターテインメントの需要から多数のウェアラブルロボットが開発されてきた. 
  これらは人間の腕や脚を補助する身体強化型から余剰の腕をつける身体拡張型まで様々であり, 本研究ではその中でも身体拡張型のウェアラブルロボットに着目する. 
  一方で, それらは人間の四肢の能力に匹敵するほどの力と動作範囲を確保できておらず, エンターテインメントや軽い物体のマニピュレーションタスクへの適用に限られている. 
  そこで本研究では, 人間を持ち上げることができる程の力を持ちつつ, マニピュレーションまでを可能とした身体拡張型ウェアラブルロボットVlimbを開発する. 
  これは腱駆動機構の利点を活かしたワイヤ経由点の変更機能により, 繊細な動きを伴うマニピュレーションと力強さを必要とする人持ち上げタスクを両立する. 
  また, 可動範囲に問題点のある腱駆動機構の欠点を受動リング構造の導入により解決することで, 人間と同様の力強さと可動域を備えている. 
  本論文ではこのVlimbの設計法と予備的なマニピュレーション・人持ち上げタスクを行い, その有効性を確認するとともに, 現状の機構の問題点について考察した.
  }%
\end{abstract}

\section{Introduction} \label{sec:introduction}
\switchlanguage%
{%
In recent years, wearable robots have been utilized in various applications such as entertainment and physical assistance. 
In medical contexts, the robot suit HAL \cite{Hayashi2015HAL, CYBERDYNEHAL} is employed for rehabilitation by assisting the movements of human lower limbs. 
In agricultural settings and similar applications, devices like the Muscle Suit \cite{MusclesSuits} are utilized to support human upper limbs, enabling individuals to lift heavier loads more easily than expected.
Within the realm of entertainment, there exist devices known as ``Jizai Arm'' \cite{Yamamura2023JizaiArm} which enable new artistic expressions by attaching additional limbs to dancers, allowing them to perform dances with unconventional limb configurations.
The development of wearable robots for diverse purposes is evident across society. 

Presently, research is developing body enhancing robots called Supernumerary Robotic Limbs (SRL) \cite{yang2021supernumerary}.
This technology involves attaching robotic arms or similar appendages to the human body, thereby extending its capabilities and enabling movements previously considered impossible.
Parietti et al. have developed SRL with three degrees of freedom, capable of assisting in assembly tasks by maintaining body balance with a rod inserted into the ground \cite{parietti2014supernumerary}.
Additionally, SRLs have been devised to efficiently support assembly work on aircraft, where solo assembly would be challenging, by using rods protruding from the waist through various supporting methods \cite{parietti2016supernumerary}.
Furthermore, initiatives such as Metalimbs, developed by Sasaki et al., remap foot movements to robotic arms attached to the waist, enhancing versatility in desk work \cite{sasaki2017metalimbs}. 
Aizono et al. have developed a ``third arm'' \cite{amano2019development, iwasaki2022ExtraLimb} capable of assisting human life activities through speech and eye contact commands.
Additionally, SRLs with three degrees of freedom\cite{Veronneau20203-DOF}, powered externally, have been developed for tasks such as fruit harvesting and wall painting. 
\tabref{table:SRLrobots} summarized the SRL robot had been developed.

\begin{figure}[t]
    \centering
        \includegraphics[width=0.95\columnwidth]{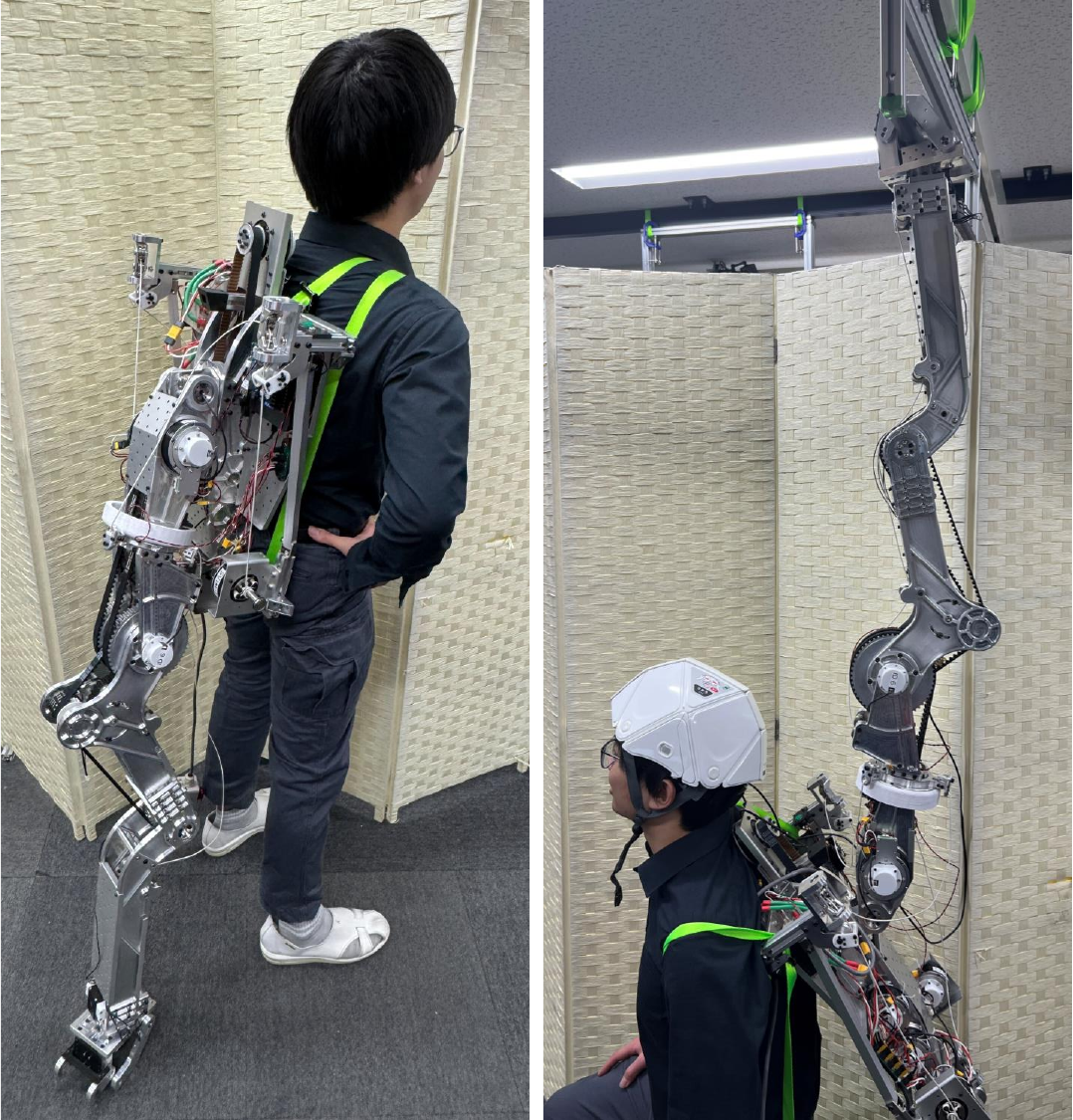}
        \caption{Vlimb attached to a human. The left figure depicts an overall view from the back, while the right one shows Vlimb gripping a fixed bar and lifting the human.}
        \label{figure:overview_with_human_EN}
\end{figure}

However, the aforementioned prior research studies have a common issue: they struggle to reconcile both the powerfulness and reachability required for human-like functionality in limbs. 
Reachability here refers to both the number of degrees of freedom (DOFs) and the length of the longest link.
Powerfulness here refers to the extent of tasks that can be performed with the maximum joint torque that the robot can exert. 
For instance, in the case of SRL \cite{parietti2014supernumerary}, while it exhibits sufficient force to maintain human body balance, its degrees of freedom are limited to three. 
On the other hand, Metalimbs \cite{sasaki2017metalimbs} offer seven degrees of freedom, but their manipulation capabilities are restricted to lightweight objects such as mobile phones or soldering irons.
\begin{table}[tb]
  \centering
  \vspace{2.0ex}
  \caption{Physical parameters of Vlimb}
  \label{table:SRLrobots}
  \small
  \begin{tabular}{lcr} \toprule
    Robot & DOFs / Length & Task \\ \midrule
    SRL2014  \cite{parietti2014supernumerary} & 4 / \SI{1.46}{\metre} & Balacing users \\ 
    Metalimbs \cite{sasaki2017metalimbs} & 7 / \SI{0.7}{\metre} & Manipulating \\
    Iwasaki arm \cite{amano2019development, iwasaki2022ExtraLimb} & 4 / \SI{0.7}{\metre} & Manipulating \\
    Veronneau arm\cite{Veronneau20203-DOF} & 3 / \SI{1.3}{\metre} & Manipulating \\ \bottomrule
  \end{tabular}
  \vspace{2.0ex}
\end{table}

The challenge stems from the inherent difficulty in achieving a balance between powerfulness and a wide reachability in wearable robots where lightweightness is desired. 
Pursuing greater powerfulness necessitates a high gear reduction ratio, resulting in increased weight of the gear mechanism. 
Furthermore, enhancing the mechanism's powerfulness to withstand exerted forces leads to further weight gain. 
Seeking a wider reachability requires more degrees of freedom and longer linkages, ultimately increasing the weight due to additional joint mechanisms and material.
Therefore, this research endeavors to develop Vlimb, a wearable robotic system capable of lifting humans while also enabling manipulation tasks, 
for example, retrieving an object from a location that would normally be out of reach for a human hand. 
By addressing the challenges of powerfulness and reachability, Vlimb aims to offer a comprehensive solution for enhancing human capabilities through wearable robotics.

In the development process, this study focused on the characteristics of wire routing mechanisms. In wire-driven systems, the arrangement of wires significantly affects the characteristics of the same joint structure. 
By employing different wire configurations for different tasks, it becomes possible to accommodate both delicate manipulations requiring fine movements and tasks involving lifting humans that demand powerfulness, all within the same mechanism.
Additionally, there are issues with the reachability in tendon-driven mechanisms. 
In systems such as FALCON \cite{kawamura1997development} and the tensegrity-inspired compliant three degree-of-freedom robotic joint developed by Friesen et al. \cite{Friesen2018ICRA}, when rotation is applied in a direction perpendicular to the axis along which the wire is tensioned, a phenomenon occurs where the wire wraps around the link.
Current solutions attempt to mitigate this by positioning the fixed points of the wire as far as possible from the link, allowing some leeway before wrapping occurs.
However, in wearable robots where design space is constrained, such designs are challenging.
As a solution, this research proposes a passive ring structure where the fixed points of the wires are attached to movable rings. 
This approach allows for a reachability comparable to that of humans while preventing the wrapping issue. 
Incorporating these design elements, Vlimb was developed.

The contributions of this research can be summarized as follows:
\begin{itemize}
  \item Achieving a balance between powerfulness and dexterity in wearable robotic systems through the alteration of wire routing mechanisms in bodily extension wearable robots.
  \item Proposing a passive ring structure to address the problem of ensuring a wide reachability in wire-driven systems, which is typically challenging to achieve.
  \item Designing and developing Vlimb, a bodily extension wearable robot incorporating the aforementioned features, and validating its effectiveness through preliminary experiments.
\end{itemize}
}%
{%
近年ウェアラブルロボットはエンターティメントや身体補助などで使用されている．
医療用などの用途では、ロボットスーツHAL\cite{Hayashi2015HAL}が使用されている．人間の下肢の運動アシストを行うことで，リハビリテーションを行える．
農業用などに用途では，マッスルスーツなどを使用される．人間の上肢を支えることによって，想定より重いものをより軽く持ち上げらることが可能である．
エンターテイメントにおいては自在肢\cite{Yamamura2023JizaiArm}と呼ばれるものが作成されている．
ダンサーに本来存在していない四肢を取り付けダンスを行うことにより芸術面において新しい表現を行っている．
このように社会ではウェラブルロボットが様々な用途で開発されている．

現在研究ではSupernumerary Robotic Limbs(SRL)と呼ばれる身体拡張ロボットが開発されている．\cite{yang2021supernumerary}
これは，人間の身体にロボットアームなどを取り付けることにより，自身の身体の機能を拡張し，今までの身体では不可能のような動きが可能となる．
Pariettiらが開発したSRLは，3自由度の棒を器用に使い，身体バランスを維持可能にすることで，
組み立て作業の補助を行うもの\cite{parietti2014supernumerary}が開発されている．
また，一人では組み立てが困難な航空機の組み立て作業に対して，
腰から生えている棒を様々な手法を用いて支えることで効率的に作業を行えるようになるSRL\cite{parietti2016supernumerary}が開発されている．
また，佐々木らが足の動きを，腰から生えているロボットアームにリマッピングすることで，机作業を多彩にするMetalimbs\cite{sasaki2017metalimbs}が開発されたり，
天野らが発話やアイコンタクトで動かせる人間の生活を手助けが可能となる第三の腕\cite{iwasaki2022ExtraLimb}\cite{amano2019development}が開発されたり
外部に動力源を持つことで，果物の収穫や壁の塗り替えなどが行える三自由度のSRL\cite{Véronneau20203-DOF}などが開発されている．
\tabref{table:vlimbparams}に各ロボットについての仕様をまとめた.

\begin{table}[tb]
  \centering
  \vspace{2.0ex}
  \caption{Physical parameters of Vlimb}
  \label{table:vlimbparams}
  \small
  \begin{tabular}{lcr} \toprule
    Robot & DOF & Overall link length & Design functions\\ \midrule
    SRL2014 Parietti et al. \cite{parietti2014supernumerary} & 4 & \SI{0.5}{\metre} \\ 
    SRL2016 Parietti et al. \cite{parietti2016supernumerary} & 4 & \SI{1}{\metre} \\
    Metalimbs \cite{sasaki2017metalimbs} & 7 & \SI{0.7}{\metre} \\
    third arm iwasaki\cite{iwasaki2022ExtraLimb}\cite{amano2019development} & 4 & \SI{0.7}{\metre} \\
    third arm Véronneau\cite{Véronneau20203-DOF} & 3 & \SI{0.5}{\metre} \\\bottomrule
  \end{tabular}
  \vspace{2.0ex}
\end{table}

\begin{figure}[t]
    \centering
        \includegraphics[width=0.9\columnwidth]{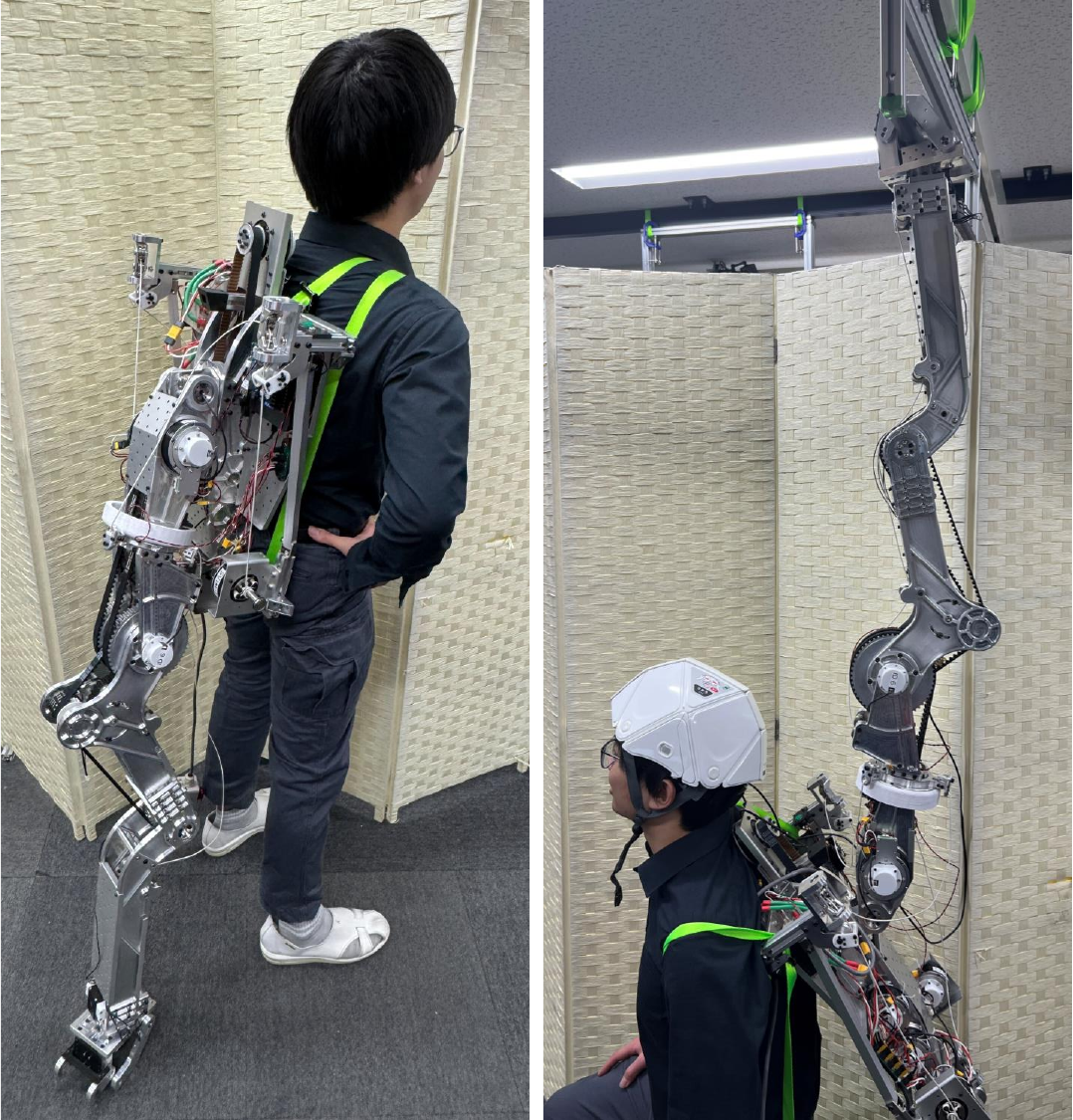}
        \caption{Vlimb attached to a human. The left figure depicts an overall view from the back, while the right one shows Vlimb gripping a fixed bar and lifting the human.}
        \label{figure:overview_with_human}
\end{figure}

しかし，これらの言及した先行研究は元来人間が持つ四肢の機能に対して，力と動作範囲を両立できていないという問題点がある．
例として，SRL\cite{parietti2014supernumerary}では人の身体バランスを保てるほどの発揮力があるが，その自由度は3にとどまる．
一方，metalimbs\cite{sasaki2017metalimbs}は自由度は7であるが，マニピュレーション対象は携帯電話や半田ごてといった軽量物に限られている．

これが達成できていないのは，軽量さが求められるウェアラブルロボットにおいて力強さと広い動作範囲の両立は困難であるからだ。
力強さを追い求めると，高減速比が必要となり，減速機の重量が増加する．
さらには，発揮する力に耐えるために機構の強度を上げる必要となり，さらに重量が増加する．
動作範囲を求めると，自由度をより多く，リンクの長さをより長くする必要があり，結果として，関節の機構や素材の重みが増していく．

そこで本研究では, 人間を持ち上げることができる程の力を持ちつつ, マニピュレーションまでを可能とした身体拡張型ウェアラブルロボットVlimbを開発する. 

開発にあたり本研究では，腱駆動機構の経由点の特性に着目した．
ワイヤ駆動においてワイヤをどのように配置するかによって，同じ関節構造であってもその特性が大きく変化する．
異なるタスクに対し，それぞれ異なるワイヤ配置にすることで，同一の機構で繊細な動きを伴うマニピュレーションと力強さを必要とする人持ち上げタスクを両立する. 

また, 腱駆動機構には可動範囲に問題点のある．
FALCON\cite{kawamura1997development}やFriesenらが開発したTensegrity-Inspired Compliant Three degree-of-freedom(DOF) robotic joint\cite{Friesen2018ICRA}
などではワイヤが張っている長軸方向に対して，垂直となる方向に回転を行うと，リンクにワイヤが巻き付くような現象が確認される。
これに対して，現状の解決策ではできる限りワイヤを固定する点をリンクから遠ざけることにより，巻き付くまでの猶予を稼いでいる。
本研究ではその解決方法として，ワイヤの固定点を可動式のリング状に固定する受動リング構造を提案する．
これにより, 人間と同様の可動域を備えることができるようになる
以上の設計要素を含むVlimbを作成した．

本研究の貢献としては以下の三点があげられる
\begin{itemize}
  \item 身体拡張ウェアラブルロボットにおける，ワイヤ経由点変更による力強さと器用さの両立
  \item 可動域確保しにくいワイヤ駆動の問題点を解決する受動リング構造の提案
  \item 以上二つの特徴を持つ身体拡張ウェアラブルロボットVlimbの設計開発と予備実験による有効性の検証
\end{itemize}
}%

\section{Design of Wire-driven Wearable Robot: Vlimb} \label{sec:method}
\switchlanguage%
{%

In \secref{sec:introduction}, the design requirements for achieving powerfulness and wide reachability in a wearable robot can be summarized as follows:
\begin{itemize}
\item Lightweight enough to be carried by a human.
\item Possessing a reachability similar to that of human limbs.
\item Strong enough to lift humans.
\end{itemize}
Based on these criteria, we developed a wire-driven bodily extension wearable robot named Vlimb. 
The name ``Vlimb'' signifies its resemblance to human limbs and its role as a fifth limb. 
The specifications of Vlimb are summarized in \tabref{table:vlimbparams}.
In the following, we propose a design method to satisfy the three elements listed above.

\begin{figure}[tb]
    \centering
        \includegraphics[width=0.95\columnwidth]{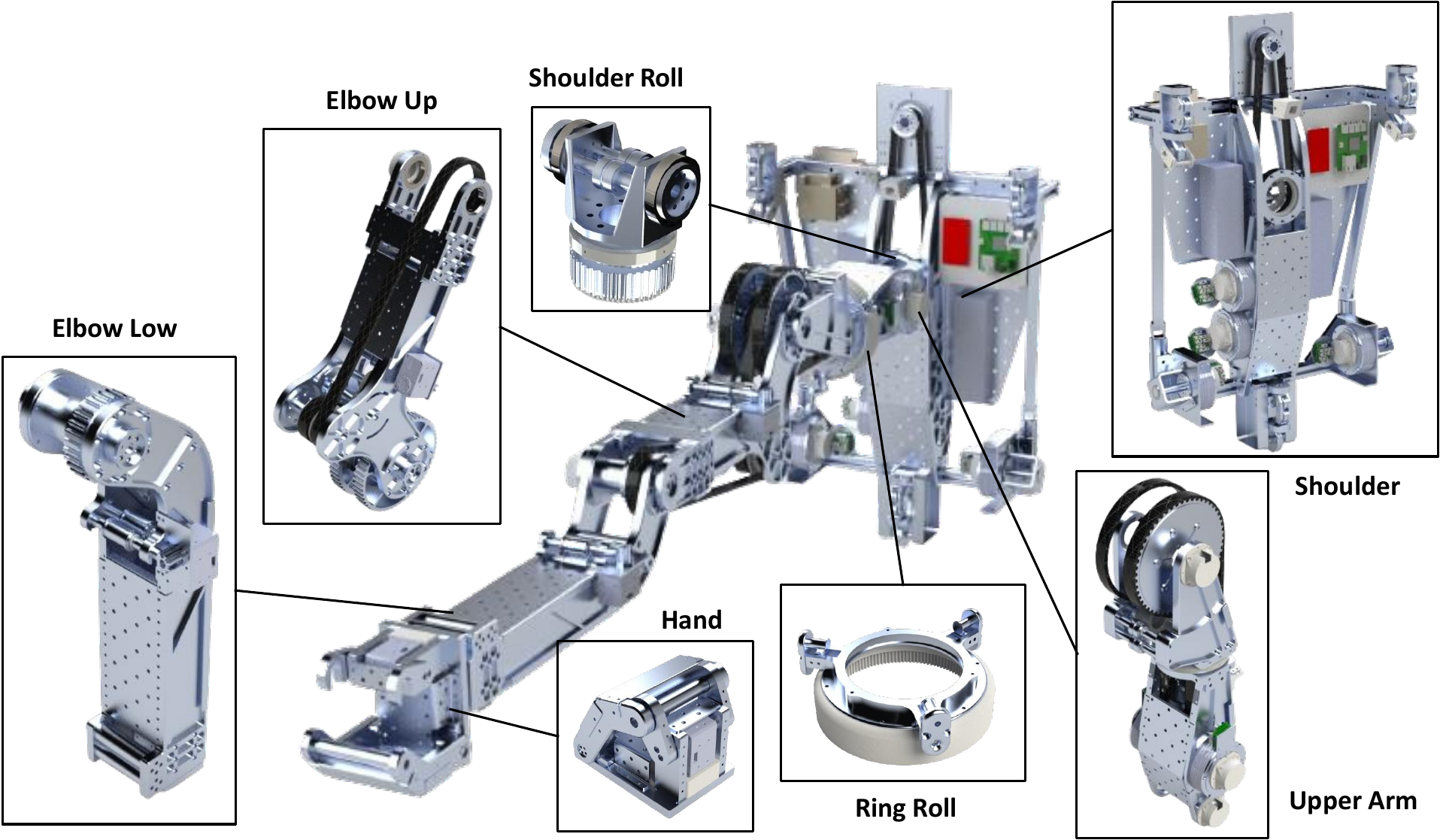}
        \caption{The overall mechanism of Vlimb. Vlimb has 5 degrees of freedom, with the link configuration arranged sequentially from the Shoulder, Shoulder Roll, Upper Arm, Elbow Upper, Elbow Lower, and Hand.}
        \label{figure:Detail_design_of_Vlimb}
\end{figure}

\begin{table}[tb]
  \centering
  \vspace{2.0ex}
  \caption{Physical parameters of Vlimb}
  \label{table:vlimbparams}
  \small
  \begin{tabular}{lcr} \toprule
    Parameter & Value \\ \midrule
    Overall height  & \SI{0.5}{\metre} \\ 
    Overall width & \SI{0.4}{\metre} \\
    Overall link length & \SI{1.3}{\metre} \\
    DOFs & 5 \\
    Total mass & \SI{16.3}{\kilogram} \\
    ShoulderRoll joint movable range& \SI{-3.14}{\radian} to \SI{+3.14}{\radian}\\
    UpperArmPitch joint movable range& \SI{-1.3}{\radian} to \SI{+1.3}{\radian}\\
    ElbowUpPitch joint movable range& \SI{-1.57}{\radian} to \SI{+1.8}{\radian}\\
    ElbowLowPitch joint movable range& \SI{-0.8}{\radian} to \SI{+2.8}{\radian}\\
    WristRoll joint movable range& \SI{-3.14}{\radian} to \SI{+3.14}{\radian}\\ \bottomrule
  \end{tabular}
  \vspace{2.0ex}
\end{table}

\subsection{Lightweight Design with Wire-Driven Mechanism}

Vlimb, developed in this study, adopts a wire-driven mechanism. 
The wire-driven mechanism of Vlimb consists of three main components: Pulley Sections for winding the wire, Waypoint sections to define the path of the wire within the mechanism, and End sections for exerting force, as illustrated in \figref{figure:wire_driven_overview}.

By utilizing wire-driven mechanisms, Vlimb benefits from the following advantages:
\begin{itemize}
    \item Achieving high gear reduction ratio while maintaining lightweight
    \item Compact motor placement
    \item High backdrivability and flexibility in environmental interaction
\end{itemize}

\begin{figure}[tb]
    \centering
        \includegraphics[width=0.95\columnwidth]{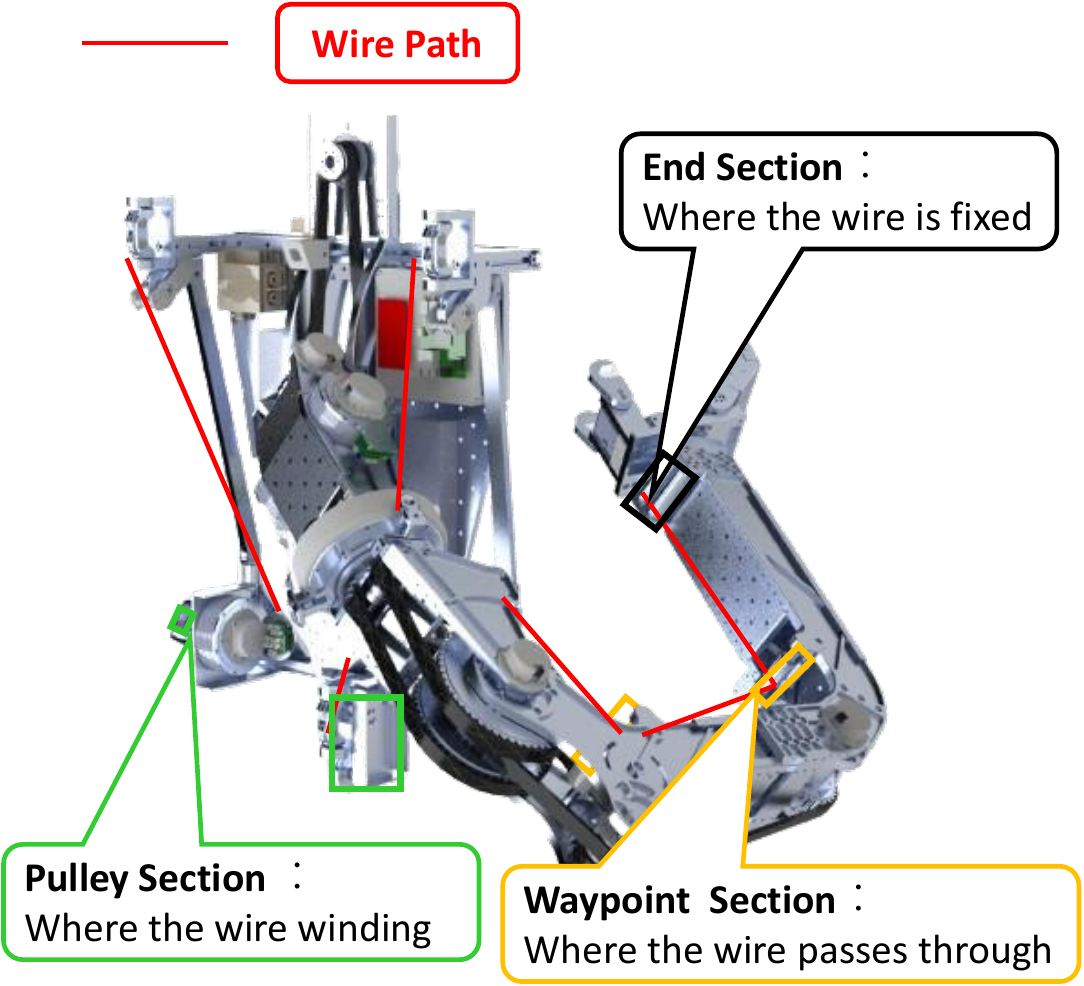}
        \caption{The components of wire-driven actuation, consisting of Pully Section, Waypoint Section, and End Section. The red lines in the figure show the wire arrangement in Vlimb.}
        \label{figure:wire_driven_overview}
\end{figure}

To further reduce weight, aluminum was chosen as the material for Vlimb, and additional weight reduction was achieved through machining processes such as milling and pocketing.

Furthermore, it is essential to aggregate actuators at the root link. 
Therefore, considering the ease of mechanism design, we opted to design using belts in places where they can be substituted. 
As a result, the weight of Vlimb was reduced to 16.3 kg, making it feasible for attachment to humans.

\subsection{Advantages of Changing Wire Waypoints for Balancing powerfulness and reachability}

In wire-driven robots, not only the configuration of the links but also the placement of wire waypoints significantly affects the mechanism's powerfulness and reachability.

The challenge in wearable robotics is to achieve a balance between powerfulness and reachability within the same device. 
As mentioned in \secref{sec:introduction}, when simultaneously achieving powerfulness and reachability, the weight of the device becomes too heavy to be carried by a human. 
To address this, we propose a method of transitioning between two modes using wire waypoint changes.

As illustrated in \figref{figure:keiyuten_hennkou_merit}, this method implements two modes: a Power Mode for achieving powerfulness and a Manipulation Mode with a wide reachability. 
While the mechanical structure of the device remains the same, the difference lies in whether the wire passes through the waypoints. 
Consequently, the achievable joint torque and degrees of freedom vary. In Power Mode, the moment arm for wire-driven joints is over five times longer compared to Manipulation Mode. However, from the perspective of degrees of freedom, two pitch degrees of freedom are replaced by linear motion, resulting in a reduction of one degree of freedom.

By implementing such a mechanism, when high torque is required to move a person, the device can switch to Power Mode by bypassing the waypoints. 
Conversely, when a wide reachability is needed for object manipulation, the device can switch to Manipulation Mode by attaching wires to the waypoints, thereby increasing the degrees of freedom.
\begin{figure}[tb]
    \centering
        \includegraphics[width=0.95\columnwidth]{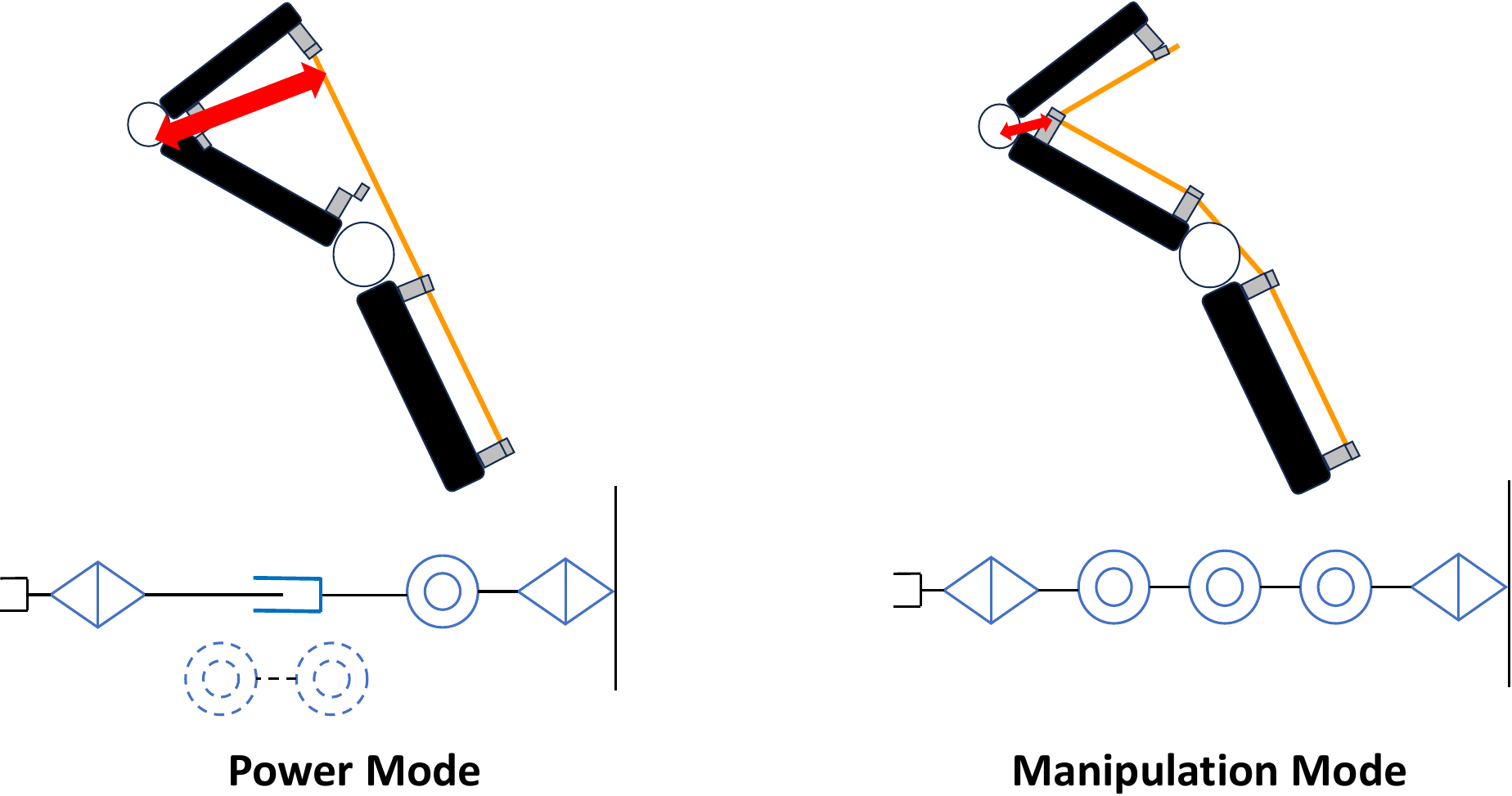}
        \caption{The difference between Power Mode and Manipulation Mode. The diagram shows that in Power Mode, a larger moment arm can be achieved compared to Manipulation Mode when considering joint torque. In comparison to Manipulation Mode, Power Mode has one less degree of freedom and becomes a translational joint.}
        \label{figure:keiyuten_hennkou_merit}
\end{figure}

\begin{figure}[tb]
  \centering
      \includegraphics[width=0.95\columnwidth]{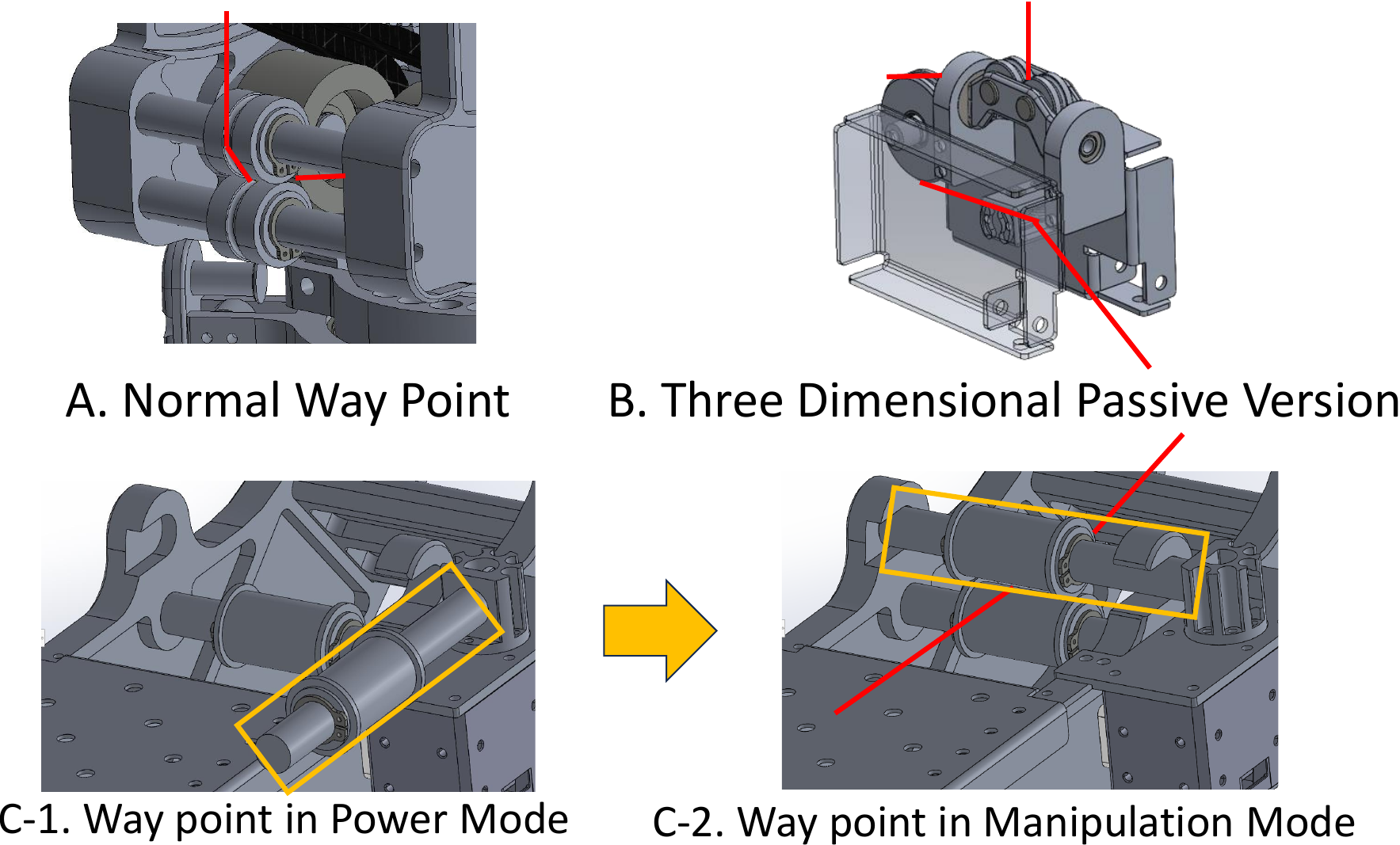}
      \caption{The design of waypoints. In A and B, wires are fixed to the waypoints, allowing passive movement in two or three dimensions. Conversely, in C, by removing the bearings, the decision to pass through the waypoints can be determined. When removed as in C-1, a structure that does not pass through the waypoints in Power Mode can be obtained, while when closed as in C-2, it becomes Manipulation Mode.}
      \label{figure:keiyuten_hennkou_overview}
\end{figure}

Vlimb has three types of wire waypoints: (A) and (B) are fixed waypoints, while (C) is a switchable waypoint, as illustrated in \figref{figure:keiyuten_hennkou_overview}.

(A) serves as a mechanism to anchor the wire at a specific position.
Since the wire moves in a plane relative to the fixed point, it needs to be constrained to a two-dimensional plane. 
To address this, the wire is designed to be sandwiched between two bearings, solving the challenge.
(B) utilizes a three-dimensional passive wire alignment device developed by Suzuki et al. \cite{Temma}.
As a solution to the limitation of (A), which confines movement to a plane and cannot accommodate the three-dimensional movement of End Sections and Waypoint sections, this device was particularly used for power transmission to the passive ring at the Shoulder joint.
(C) was designed to facilitate mode switching. 
Due to the attachment and detachment of wires, it has a wider width compared to (A) to accommodate slight deviations of the wires. 
(C-1) represents the state with the waypoint removed for Power Mode. 
(C-2) shows the state holding the wire for Manipulation Mode.
The components are actuated by servo motors. The upper bearing is rotated and secured in a groove on the frame.

To achieve the required lifting force for Power Mode, we selected motors accordingly. 
The force required to lift an adult male equivalent weight (60 kg) is approximately 600 N. 
Considering safety margins, we limited a maximum output force to around 1500 N.
The aluminium pulley with a diameter of 12 mm was selected as the take-up pulley on the grounds of its load capacity.
This motor is required to provide torque of 3.6 Nm and 9 Nm at its maximum output.
In this instance, a T-motor AK60-6 motor from the AK series was employed. 

Furthermore, as significant forces are exerted on the frames of each link, we constructed them using two machined aluminum parts with a thickness of 10 mm and two sheet metal parts with a thickness of 1.5 mm.
To withstand torsional deformation, the overall structure was designed to be box-shaped.
Motors, tensioners, waypoints, and other components were securely fixed to the machined aluminum parts.

\begin{figure}[tb]
  \centering
      \includegraphics[width=0.95\columnwidth]{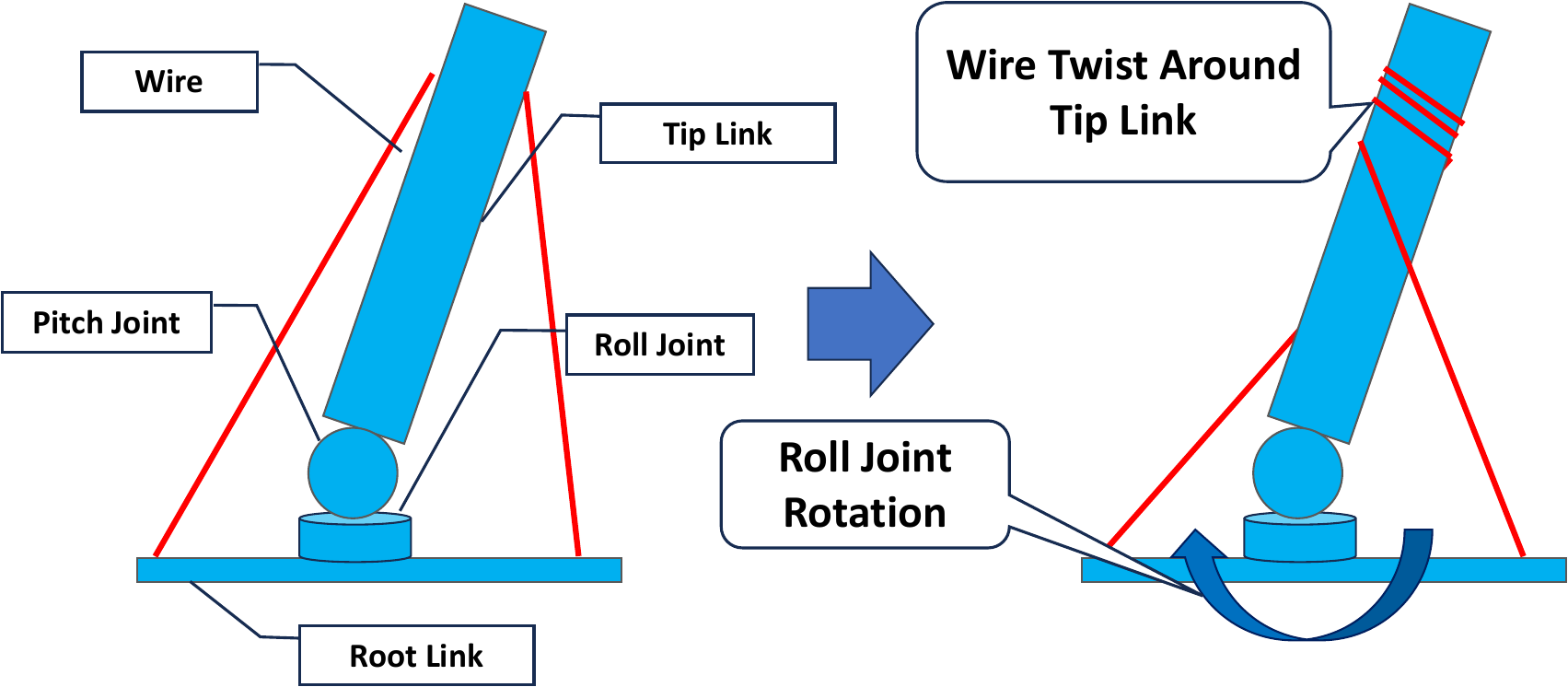}
      \caption{How the wire crossing between two links becomes twisted around the tip link due to rolling motion in the direction of rotation.}
      \label{figure:Wire-Driven_wrap_around_problem}
\end{figure}
\subsection{Ensuring reachability with Passive Ring Structure}

Vlimb adopts a passive ring structure, where the endpoints of the wires are fixed not directly to the links but to rings that move passively with bearings in between.

In conventional wire-driven robots, as depicted in \figref{figure:Wire-Driven_wrap_around_problem}, wires are often constrained along the longitudinal axis of the link. 
This can lead to a problem where, due to movements perpendicular to the direction of the wire tension, the wire wraps around the links acting as through and fixed points, as illustrated in \figref{figure:Wire-Driven_wrap_around_problem}. 
As a result, the reachability in the Roll direction, as shown in \figref{figure:Wire-Driven_wrap_around_problem}, becomes restricted.
Prior studies \cite{kawamura1997development} have addressed this issue by placing the fixed points away from the links to ensure reachability.
However, in wearable robots where design space is constrained, such designs are challenging.

To tackle this problem, we introduced the Ring Roll mechanism (see \figref{figure:Detail_of_passive_wire_fix_ring}). 
By placing bearings between the fixed point on the link and the link itself, the ring rotates passively in a way that minimizes the distance traveled by the wire under tension. 
Consequently, the bearing rotates before the wire can wrap around the link, preventing the occurrence of wrapping.
With this mechanism, Vlimb can achieve a full 360-degree reachability in the roll direction.

\begin{figure}[tb]
    \centering
        \includegraphics[width=0.95\columnwidth]{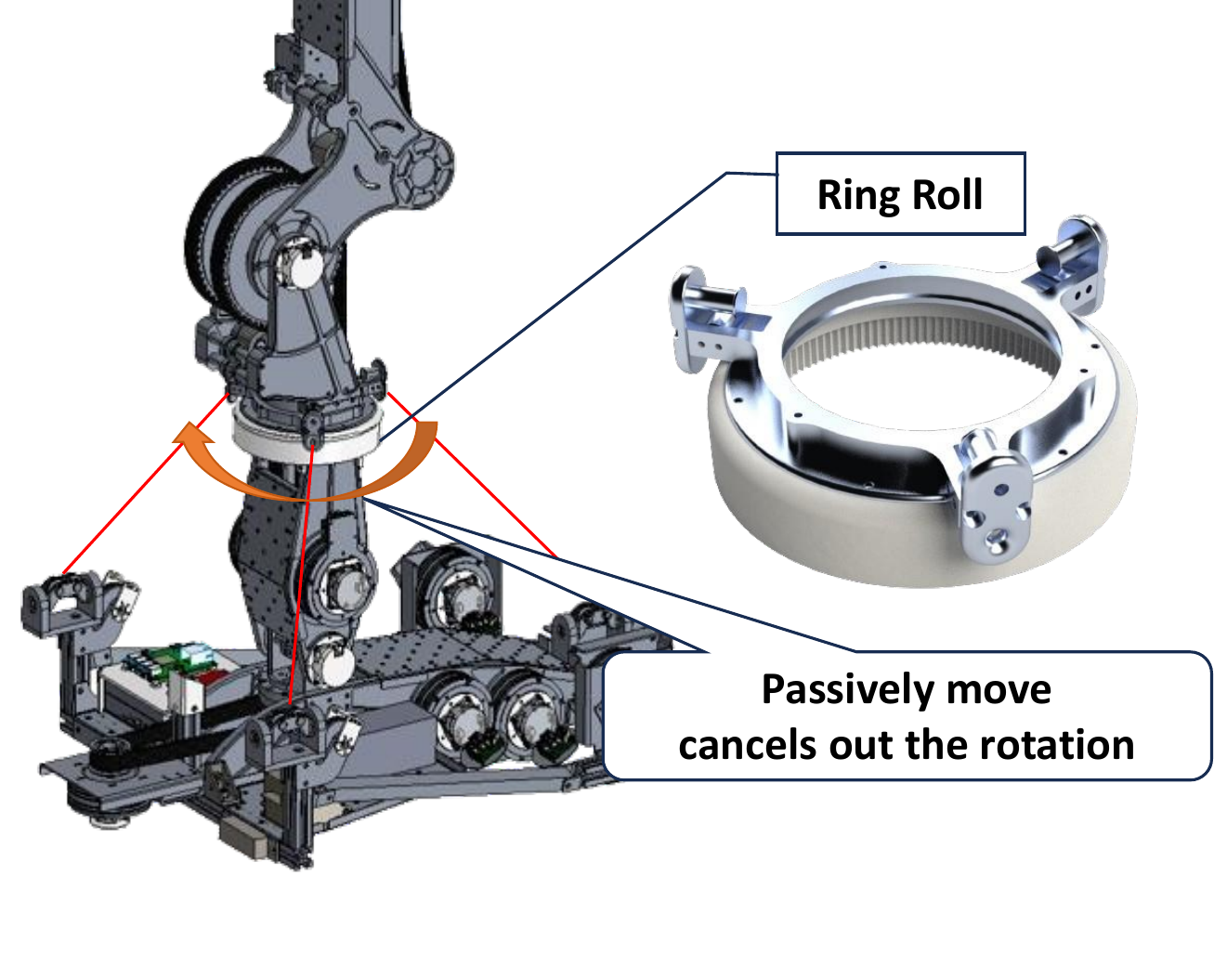}
        \caption{The mechanism of the passive ring mechanism. The red lines represent the wire routing.}
        \label{figure:Detail_of_passive_wire_fix_ring}
\end{figure}

}%
{%
\secref{sec:introduction}にて力強さと広い可動範囲をウェアラブルロボットで実現するための設計要件をまとめると，以下の三点となる
\begin{itemize}
    \item 人間が背負えるほど軽量
    \item 四肢と同じような動作範囲を持つ
    \item 人間を持ち上げられるほど力強い
\end{itemize}
これらの要素を満たすワイヤ駆動型身体拡張ウェアラブルロボットVlimbを作成した．
作成したウェラブルロボットは，人間の四肢を模し，五本目の四肢という意味を込め，Vlimbと呼称する．
Vlimbのスペックを\tabref{table:vlimbparams}にまとめた．

\begin{figure}[tb]
    \centering
        \includegraphics[width=1\columnwidth]{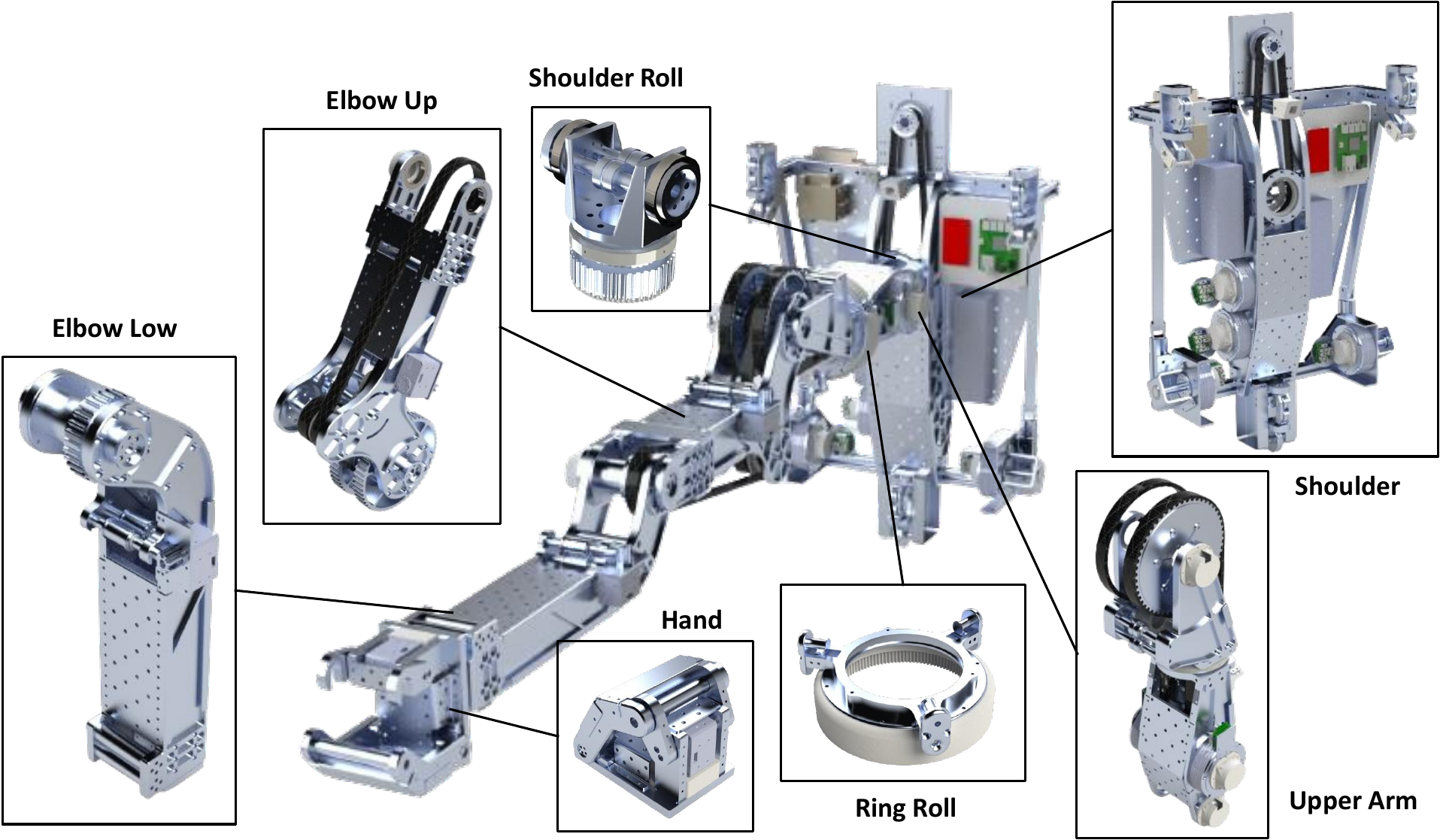}
        \caption{Overview of Vlimb}
        \label{figure:Detail_design_of_Vlimb}
\end{figure}

\begin{table}[tb]
  \centering
  \vspace{2.0ex}
  \caption{Physical parameters of Vlimb}
  \label{table:vlimbparams}
  \small
  \begin{tabular}{lcr} \toprule
    Parameter & Value \\ \midrule
    Overall height  & \SI{0.5}{\metre} \\ 
    Overall width & \SI{0.4}{\metre} \\
    Overall link length & \SI{1.3}{\metre} \\
    Total mass & \SI{16.3}{\kilogram} \\
    ShoulderRoll joint movable range& \SI{-3.14}{\radian} to \SI{+3.14}{\radian}\\
    UpperArmPitch joint movable range& \SI{-1.3}{\radian} to \SI{+1.3}{\radian}\\
    ElbowUpPitch joint movable range& \SI{-1.57}{\radian} to \SI{+1.8}{\radian}\\
    ElbowLowPitch joint movable range& \SI{-0.8}{\radian} to \SI{+2.8}{\radian}\\
    WristRoll joint movable range& \SI{-3.14}{\radian} to \SI{+3.14}{\radian}\\
    Maximum Vertical force at Manipulation Mode & \SI{30}{\newton} at \SI{15}{A}\\
    Maximum Vertical force at Power Mode & \SI{1500}{\newton} at \SI{15}{A}\\ \bottomrule
  \end{tabular}
  \vspace{2.0ex}
\end{table}

以下に上で挙げた三つの要素を満たすための設計手法を提案する．

\subsection{ワイヤ駆動を用いた軽量化}
本研究で開発したVlimbではワイヤ駆動機構を採用した．
Vlimbのワイヤ駆動機構は\figref{figure:wire_driven_overview}に示すように，ワイヤを巻き取るプーリ部，
ワイヤの機構内での経路を定めるワイヤ経由点，力を発揮する端点部で構成されている．

ワイヤ駆動を用いることにより以下の三つの利点が得られる
\begin{itemize}
  \item 軽量に高減速比を達成できる
  \item モータをルートリンク付近に集約でき，慣性モーメントを小さくすることが可能となる
  \item バックドライバビリティが高く，環境に柔軟に接触可能である
\end{itemize}
\begin{figure}[tb]
    \centering
        \includegraphics[width=0.9\columnwidth]{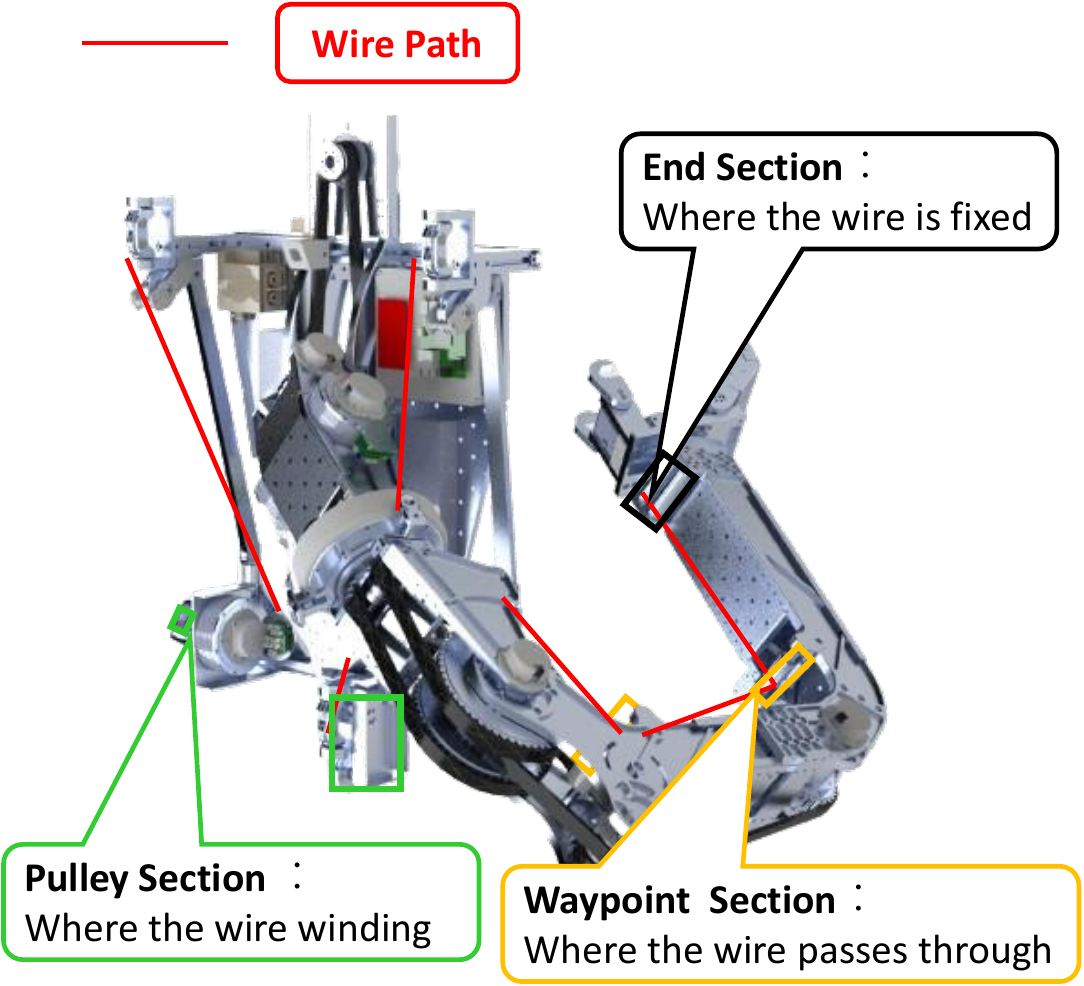}
        \caption{Wire driven overview}
        \label{figure:wire_driven_overview}

\end{figure}

軽量化のため，使用した材質はアルミであり，さらに切削による肉抜きなどを行いさらなる軽量化を行った．

また，根元のリンクに集約すれば問題ないため，機構設計の容易さを考慮し，ベルトに置き換えられる部分を置き換えた．
これによりVlimbは本体重量が16.3kgと人間に取り付けられるほどの重量となった．

\subsection{力強さと動作範囲の両立における経由点変更の利点}
ワイヤ駆動ロボットでは，リンクの構成のみならず，ワイヤの経由点の配置によって大きく機構の力強さと動作範囲が変わる．

ウェアラブルロボットにおける課題は，力強さと動作範囲の両立を同一の機体内で実現することである．
\ref{sec:introduction}章に述べたように，力強さと動作範囲を同時に達成する場合，人が背負えるほどの重量ではなくなる．
これを解決する手法として，ワイヤ駆動の経由点変更による，二つの形態の行き来を提案する．

\figref{figure:keiyuten_hennkou_merit}に示すように，本手法では力強さを達成するためのパワーモード，広い動作範囲を持つマニピュレーションモードの二つの形態を実装する．
機体の機械構造は同一のものであるが，ワイヤが経由点を通るか通らないかの違いにより，発揮できる関節トルクおよび自由度が異なる．
パワーモード時はマニピュレーションモードと比べ，ワイヤによる関節に対するモーメントアームが5倍以上になっている．
一方，自由度という観点では，二つのpitchの自由度が直動に置き換わり，自由度が一つ減っている．

このような機構を作成することにより，人を動かすほどのトルクが必要な場合はパワーモードになるよう経由点を外す．
一方，物体操作時に動作範囲が必要な場合は，マニピュレーションモードになるよう経由点にワイヤを取り付け自由度を増やすことが可能となる．

\begin{figure}[tb]
    \centering
        \includegraphics[width=0.9\columnwidth]{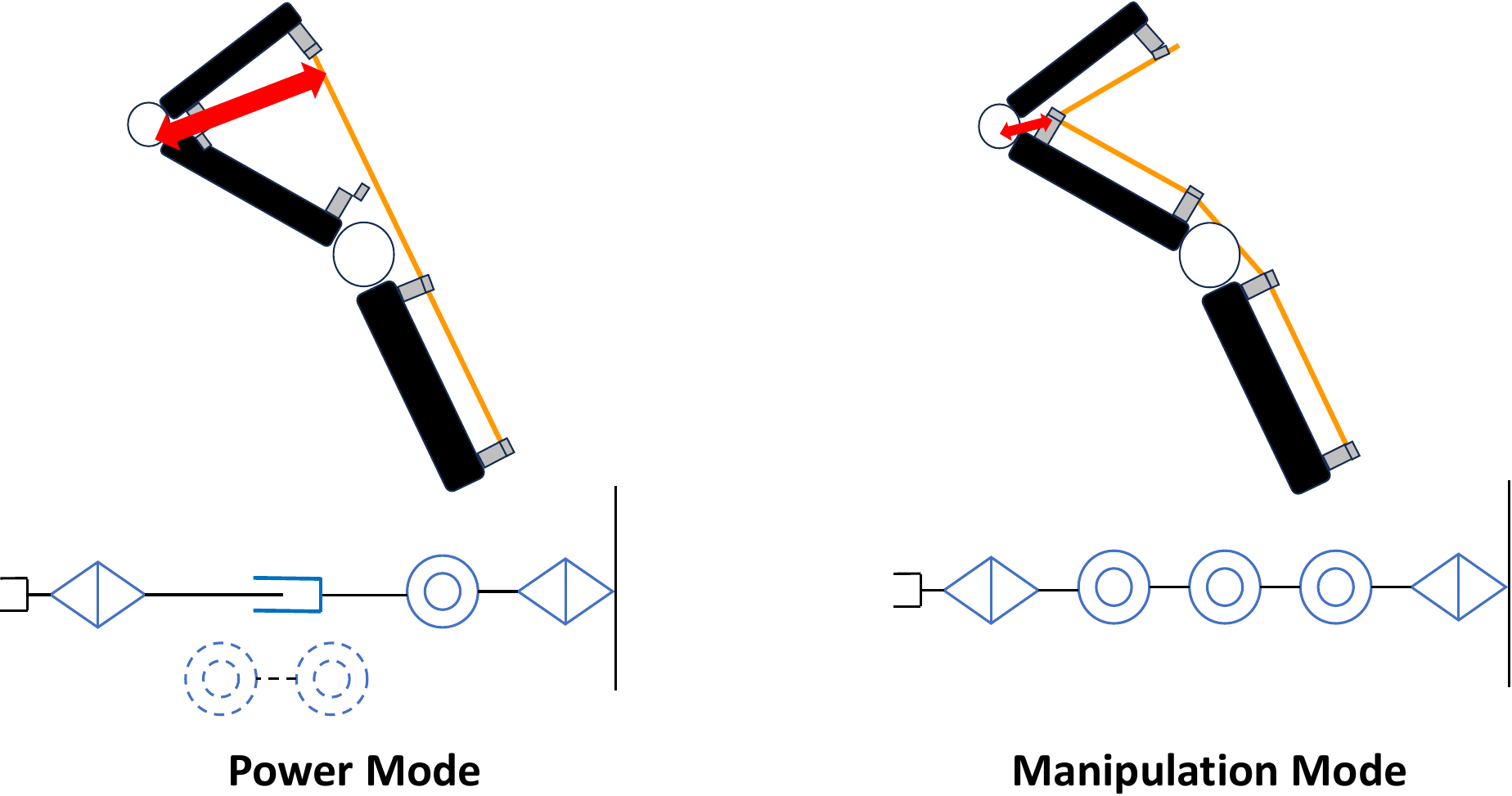}
        \caption{Difference between power mode and manipulation mode}
        \label{figure:keiyuten_hennkou_merit}
\end{figure}

Vlimbのワイヤ経由点は，三種類ある．(A)(B)は切替不能な経由点であり，(C)は切替可能な経由点．
\figref{figure:keiyuten_hennkou_overview}にて示す．

(A)はワイヤを一定の位置に留めるための機構である．
ワイヤは固定点に対し平面的に動くため，二次元平面上に固定する必要である．
それを踏まえ，このワイヤが二つのベアリングで挟まれた形状で設計することにより課題を達成している．

(B)では，本研究室の鈴木らが開発した三次元受動ワイヤ整列装置\cite{Temma}を用いた．
(A)の問題点として，平面上に拘束されており，三次元的に動く端点部，経由部に対して対応できない．
今回の設計おいては，Shoulderから受動リングへの動力伝達を行う際に三次元的にワイヤが移動してしまうため使用した．

(C)は，モード切り替えを実現するにあたって設計したものとなる．
ワイヤが付け外しする関係上，ワイヤが多少ずれても問題ないように(A)と比べ幅を大きくしている．
(C-1)がパワーモードを行うにあたって経由点を外した状態である．
(C-2)が動作範囲を確保するにあたり経由点を通すために閉めている状態である．
これらの切り替えはサーボモータを用いている．
上部にあるベアリングを回転させ、フレームの反対側の溝にロックされる仕組みとなっている．


\begin{figure}[tb]
    \centering
        \includegraphics[width=0.9\columnwidth]{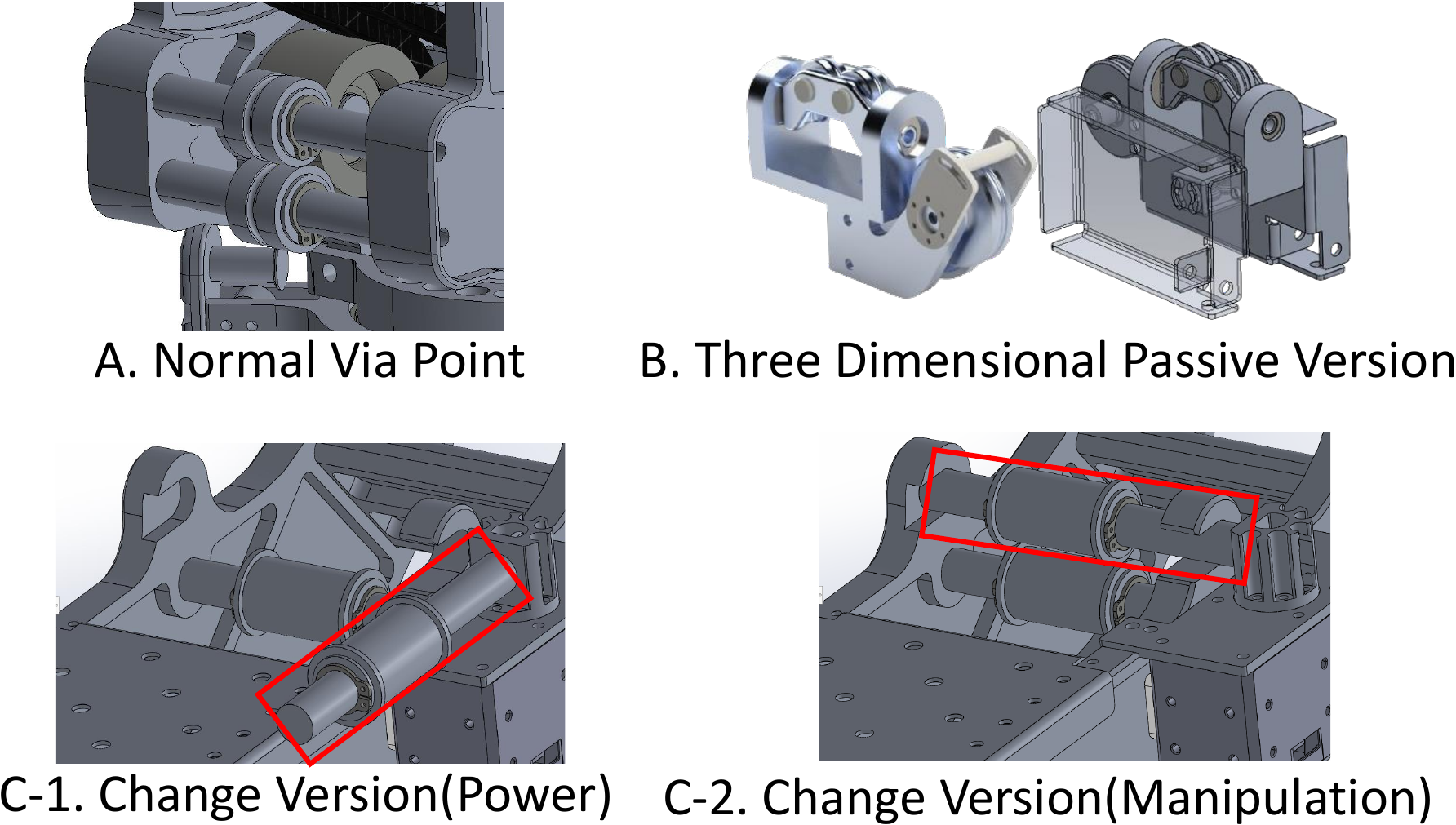}
        \caption{Design of waypoint overview}
        \label{figure:keiyuten_hennkou_overview}
\end{figure}

人を持ち上げられるほどの力をパワーモードで出せるためにモータを選定した．
成人男性相当の重量(60kg)\cite{karada}を持ち上げるために必要な発揮力は約600Nである．
安全率を考慮して，最大発揮力を1500 N程度とした．
耐荷重よりφ12のアルミ製のプーリを巻き取りプーリとして選定した．
これにより必要なトルクは3.6 Nm 最大で9 Nm発揮させる必要がある．
今回はT-motorのAKシリーズであるAK60-6というモータを使用した. 

また，各リンクのフレームに大きな力がかかるため，t10のアルミの削り出し部品を二枚とt1.5の板金二枚で構成した．
ねじり方向の変形を受けるため，全体が箱形状となるように設計した．
モータやテンショナ，経由点などをアルミの削り出し部品に固定した．

\subsection{受動リング構造による可動域の確保}
Vlimbでは受動リング構造を採用した．
これは，リンク上にワイヤの端点を固定するのではなく，ベアリングを間に挟んだ受動的に動くリングに固定する機構である．

一般的なワイヤ駆動ロボットでは\figref{figure:Wire-Driven_wrap_around_problem}にある通り，リンクの長軸方向においてワイヤの制約を受けることが多い．
本来張っているワイヤ方向と垂直方向の動きを行うことにより，ワイヤが経由部と固定部となっているリンクに巻き付いてしまう問題が発生する．
これにより，\figref{figure:Wire-Driven_wrap_around_problem}におけるRoll方向の動作範囲が狭まる．
先行研究\cite{kawamura1997development}では固定点をリンクから離して配置することにより，可動域を担保している．

この問題を，Ring Roll機構(\figref{figure:Detail_of_passive_wire_fix_ring})を用いて解決した．
リンクへの固定点と，リンクの間にベアリングを入れることでワイヤから受ける力で最も最短となるようにリングが受動的に回転する．
これにより，リンクに巻き付く前にベアリングが回転することで，巻き付きの発生を防止可能となる．
Vlimbではroll方向の可動域を360度を確保可能となった．

\begin{figure}[tb]
    \centering
        \includegraphics[width=0.9\columnwidth]{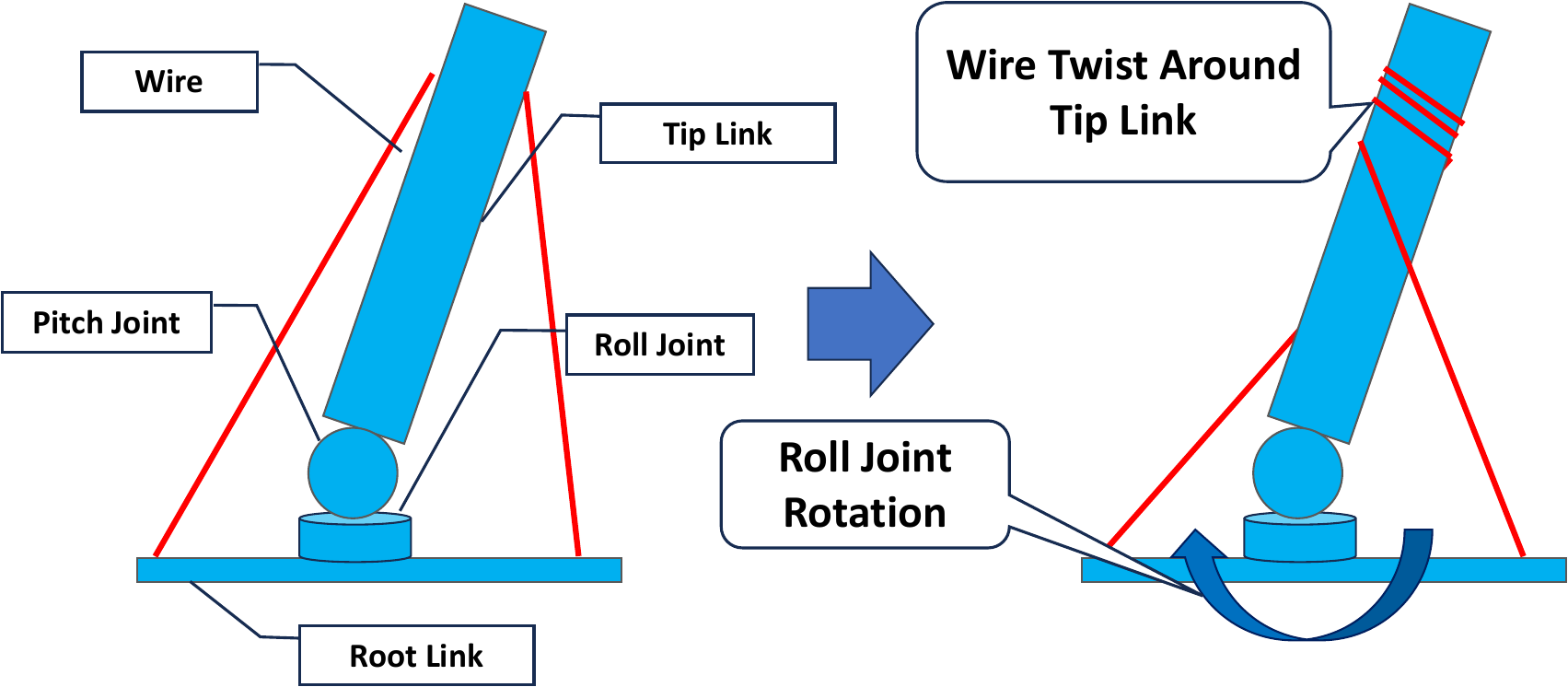}
        \caption{This illustrates how the wire crossing between two links becomes twisted around the tip link due to rolling motion in the direction of rotation.}
        \label{figure:Wire-Driven_wrap_around_problem}
\end{figure}

\begin{figure}[tb]
    \centering
        \includegraphics[width=0.9\columnwidth]{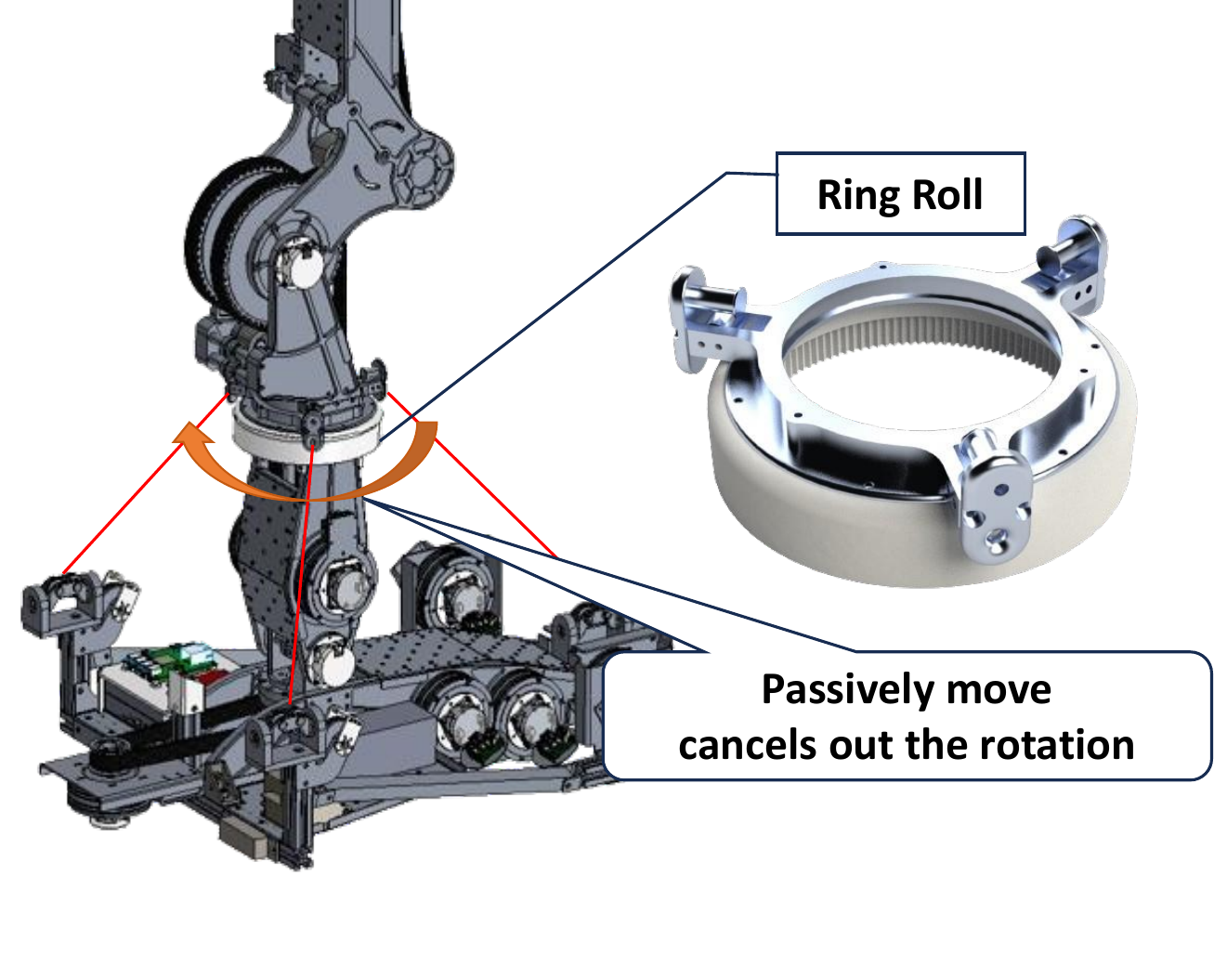}
        \caption{Detail of passive wire fix ring}
        \label{figure:Detail_of_passive_wire_fix_ring}
\end{figure}

}%

\section{System of Wire-driven Wearable Robot: Vlimb} \label{sec:system}
\switchlanguage%
{%
\begin{figure}[tb]
  \centering
      \includegraphics[width=0.95\columnwidth]{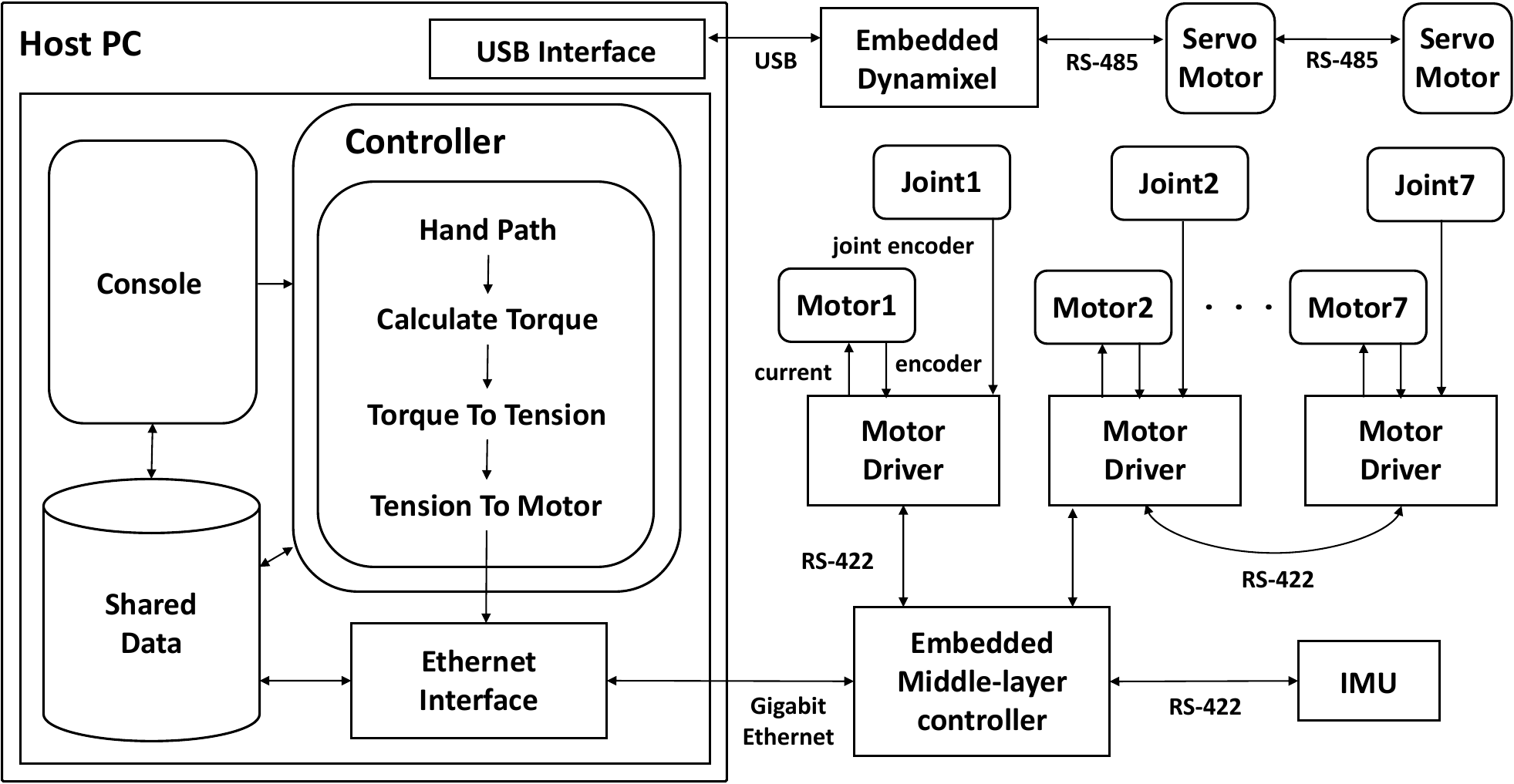}
      \caption{The overview of the Vlimb system. To drive the brushless motors, the PC has Ethernet communication with the circuit and controls seven motors via motor drivers using RS-422. On the other hand, for the actuators in the areas where the load is not large, the PC communicates with the circuit via USB to control the servo motors.}
      \label{figure:system-overview}
\end{figure}

\begin{figure}[tb]
  \centering
      \includegraphics[width=0.95\columnwidth]{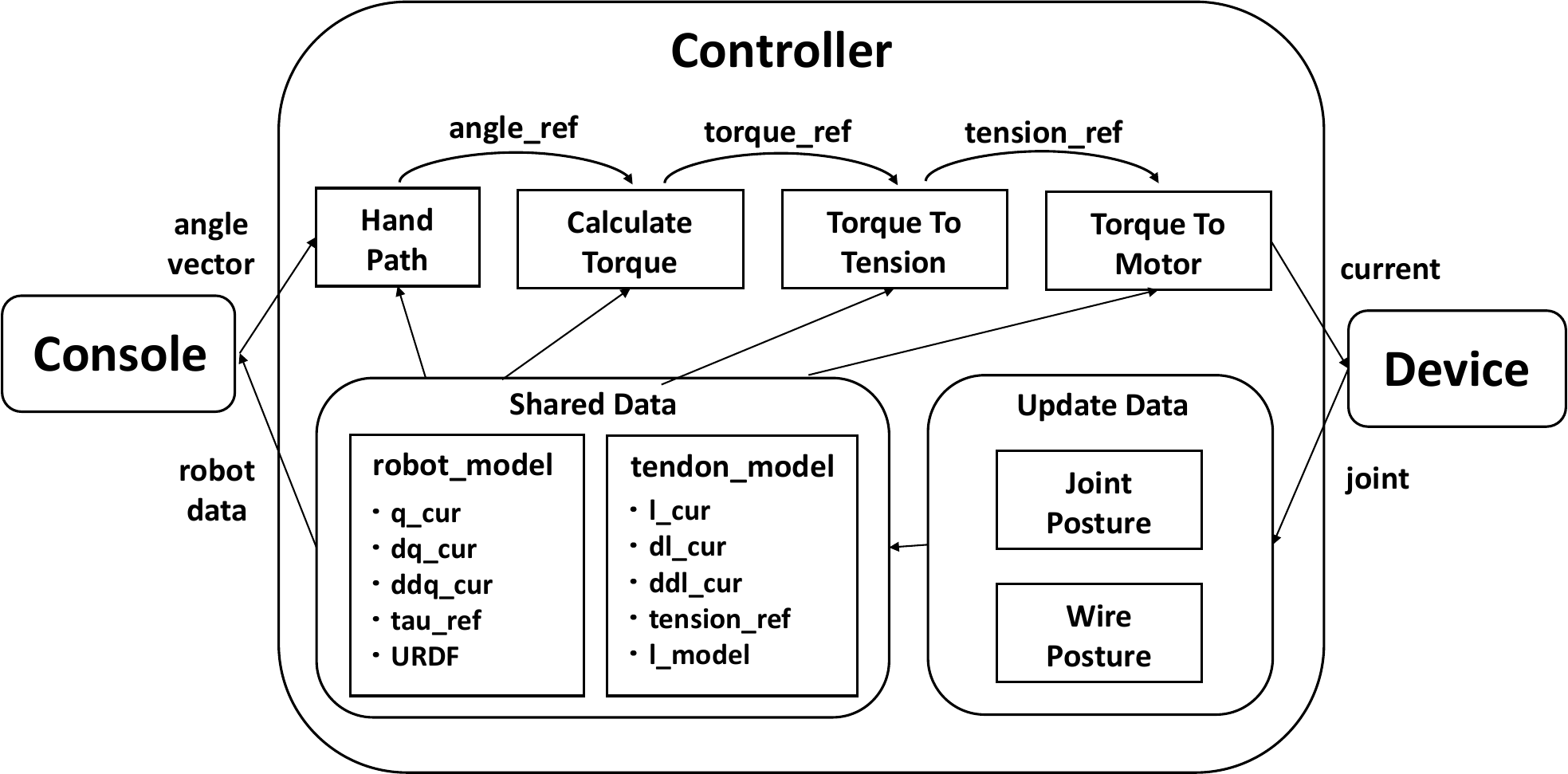}
      \caption{Diagram depicting Vlimb's controller. Humans utilize the console to read current robot information and issue angle commands. The controller processes received angle commands, converts them into current commands for output to the device. The device moves according to the received current value, transfers current joint angle information to the controller, and reflects it back to the console.}
      \label{figure:controller-overview}
\end{figure}

We have compiled the device configuration diagram of the wearable robot created in this study in \figref{figure:system-overview}.

The system comprises six brushless motors and six joint encoders. 
Each motor driver is equipped to send and receive data from the joint encoders simultaneously, facilitating precise control of each motor's position.
Due to the necessity of achieving accurate current commands, a high control frequency is required. 
To meet this demand, FPGA-based motor drivers were employed to ensure real-time motor control.

The motion of the end effector and adjustments in waypoints demand minimal power and do not require real-time responsiveness, unlike the motion of the main body. 
As a result, our focus was on minimizing power consumption. To achieve this, we opted for servo motors and operated them through a USB interface.

The following is a summary of the controller for the wearable robot created in this study in \figref{figure:controller-overview}.

The controller internally maintains angle information for each joint, as well as current wire length and tension, in common data structures called Joint Posture and Wire Posture, respectively.
Whenever this information is retrieved from the device, it is updated and then transmitted to the console for user display.
Users can input their desired joint angles through the console interface.
For each specified angle, the controller employs third-order spline interpolation to calculate the commanded angle for every corresponding time step.
Subsequently, these calculated angles are transmitted to each device for execution.

Next, the controller combines the feedforward term (gravity torque) and the feedback term (proportional control) based on the specified angles and the current joint information to compute the joint torque.
The calculation of gravity torque was performed using a library called Pinocchio \cite{carpentier2019pinocchio}.
Subsequently, the calculated joint torque is converted into the desired tension for each wire using the muscle length Jacobian \cite{Kengoro2016Jacobian}.
In Vlimb, in addition to wire-driven mechanisms, belt-driven mechanisms are also present.
By treating the belt as a derivative of wires, we have unified the controller.

Finally, the desired tension is converted into command currents for the motors, which are then passed to the motor drivers in the device layer.

}%
{%
今回の作成したウェアラブルロボットのデバイス構成図を\figref{figure:system-overview}にまとめた．

ブラシレスモータ数は6つであり，関節エンコーダ数は6つである．各モータを制御するモータドライバには関節エンコーダを同時に送受信を行えるように組み込んだ．
電流指令を正しく反映するため，高い制御周波数が必要となる．そこで，モータドライバにはFPGAを使用し，リアルタイムでモータの制御を行った．
また，研究室内の規格に合わせるためにモータモジュールを作成した．

本体のVlimbの動き以外にも手先の動きや経由点変更などで，低周期で動かしてよく，小出力なアクチュエータを利用するために、USBインターフェースを介したサーボモータも利用している．

\begin{figure}[tb]
    \centering
        \includegraphics[width=1\columnwidth]{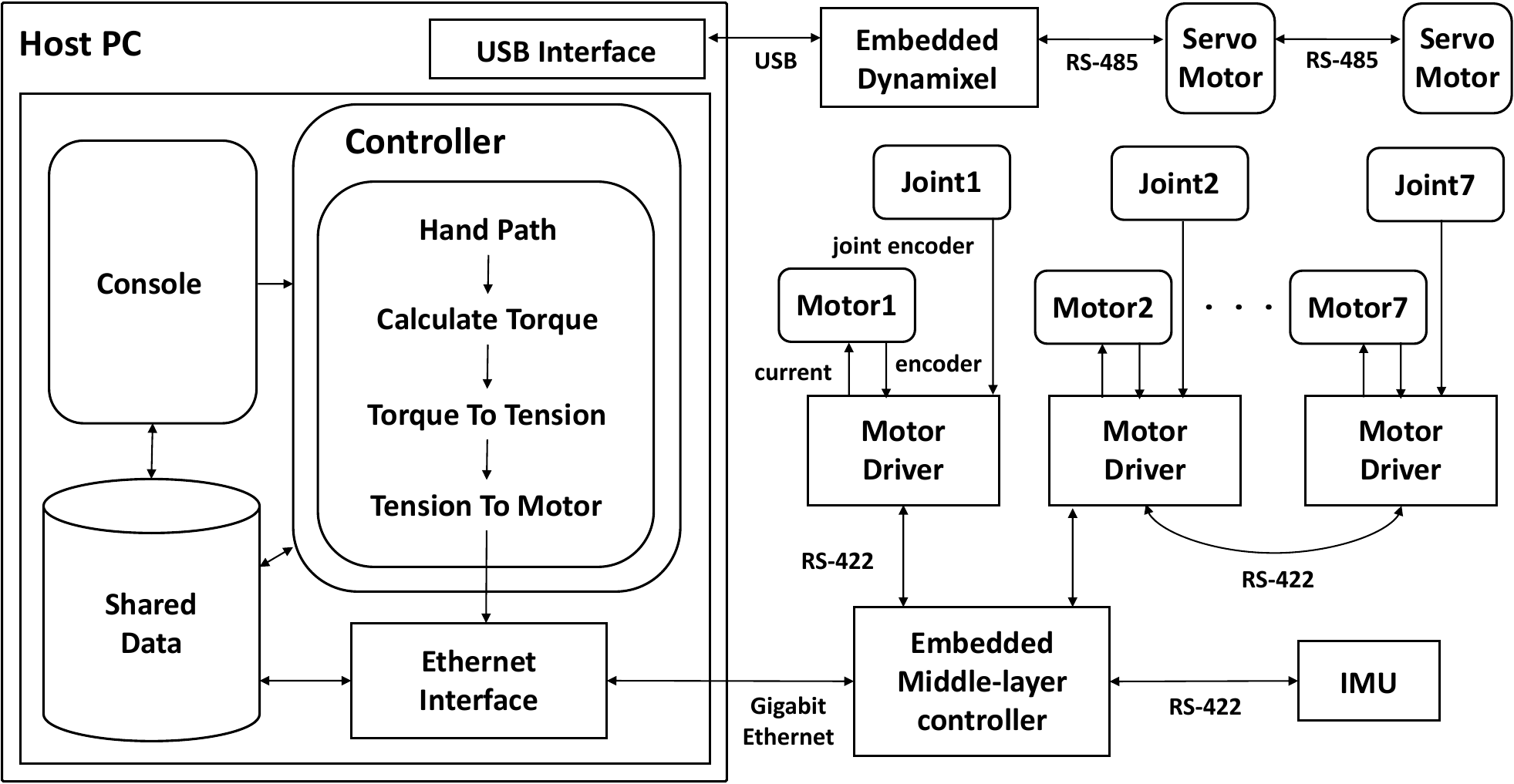}
        \caption{The overall view of the Vlimb system. To drive the brushless motors, the PC has Ethernet communication with the circuit and controls seven motors via motor drivers using RS-422. On the other hand, for the actuators in the areas where the load is not large, the PC communicates with the circuit via USB to control the servo motors.}
        \label{figure:system-overview}
\end{figure}

また今回作成したウェアラブルロボットのコントローラの概略を\figref{figure:controller-overview}にまとめた．

Jointの情報や現在のワイヤの長さ張力などのデバイス情報をJointPosture, WirePostureで更新を行い．
それらをコンソールで受け取り，人間が目標とする角度を指令する．
指令した目標角度に対して，三次スプライン補完を利用し，各タイムラインごとの指令角度を算出した．

指令された角度に対して現在の関節情報などからフィードフォワード項として重力トルクとフィードバック項として比例制御の足し合わせを行い，関節トルクを求めた．
重力トルクの算出には，pinoccio\cite{carpentier2019pinocchio}と呼ばれるライブラリを用いた．
計算し、算出された関節トルクに対して，各ワイヤへの指令張力へ変換を筋長ヤコビアンを用いることによって，各ワイヤのテンションへの変換を行った\cite{Kengoro2016Jacobian}．
ここで本研究では，拮抗ワイヤー駆動とベルトを用いた駆動がハイブリッド構造となっているが，制御器は分けず，ベルト構造に対しては押し引きが可能なワイヤとみなしている．
これにより制御器は分けず，Vlimbの制御を行うことができた．

最後に求めた張力からモータへの指令電流へと変換を行い，デバイス層のモータドライバへと渡される．

\begin{figure}[tb]
    \centering
        \includegraphics[width=1\columnwidth]{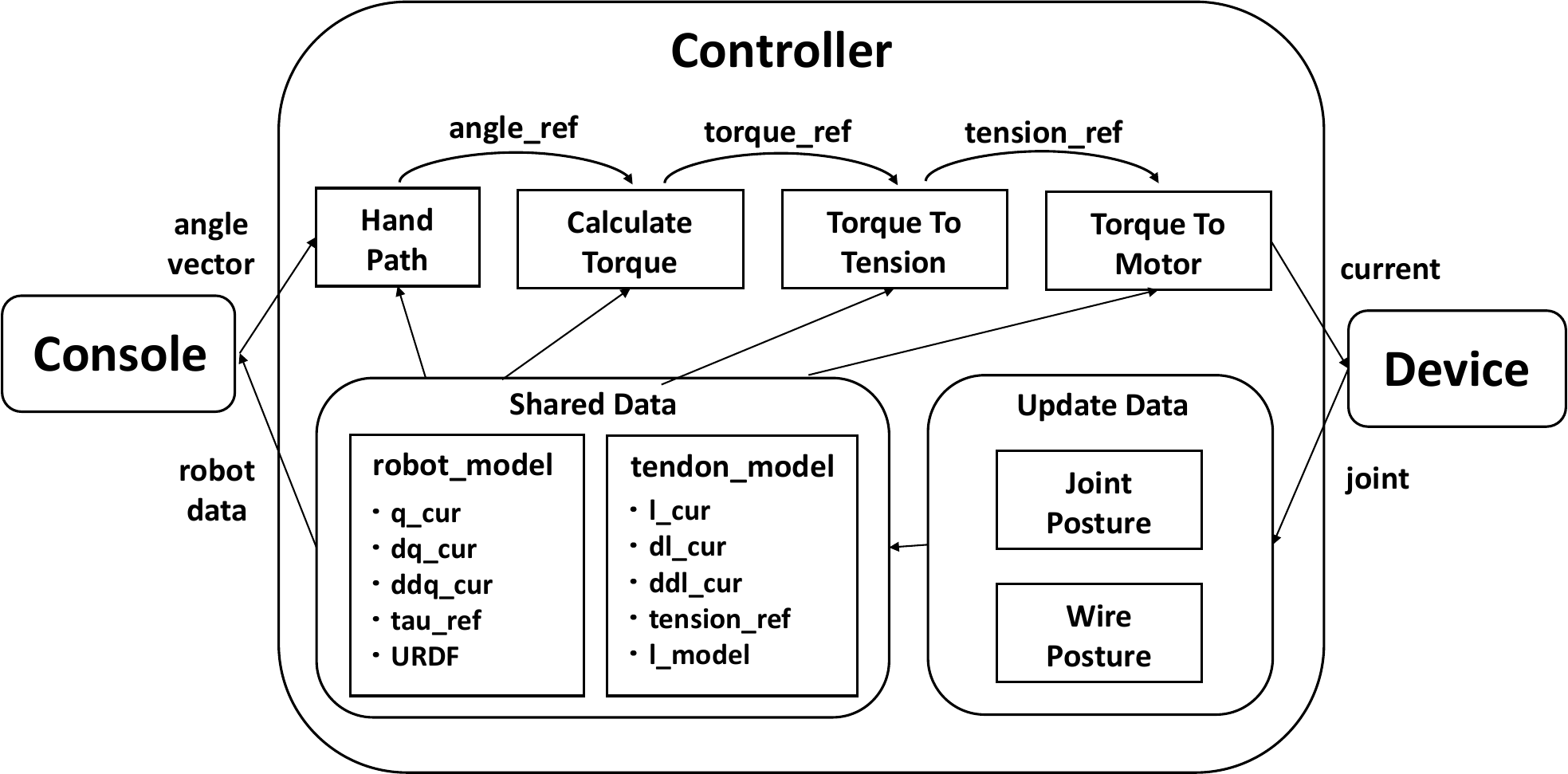}
        \caption{Diagram depicting Vlimb's controller. Humans utilize the console to read current robot information and issue angle commands. The controller processes received angle commands, converts them into current commands for output to the device. The device is operated according to the received current value, transfers current joint angle information to the controller, and reflects it back to the console.}
        \label{figure:controller-overview}
\end{figure}
}%

\section{Preliminary Experiments} \label{sec:experiments}
\switchlanguage%
{%
\begin{figure}[tb]
  \centering
      \includegraphics[width=1\columnwidth]{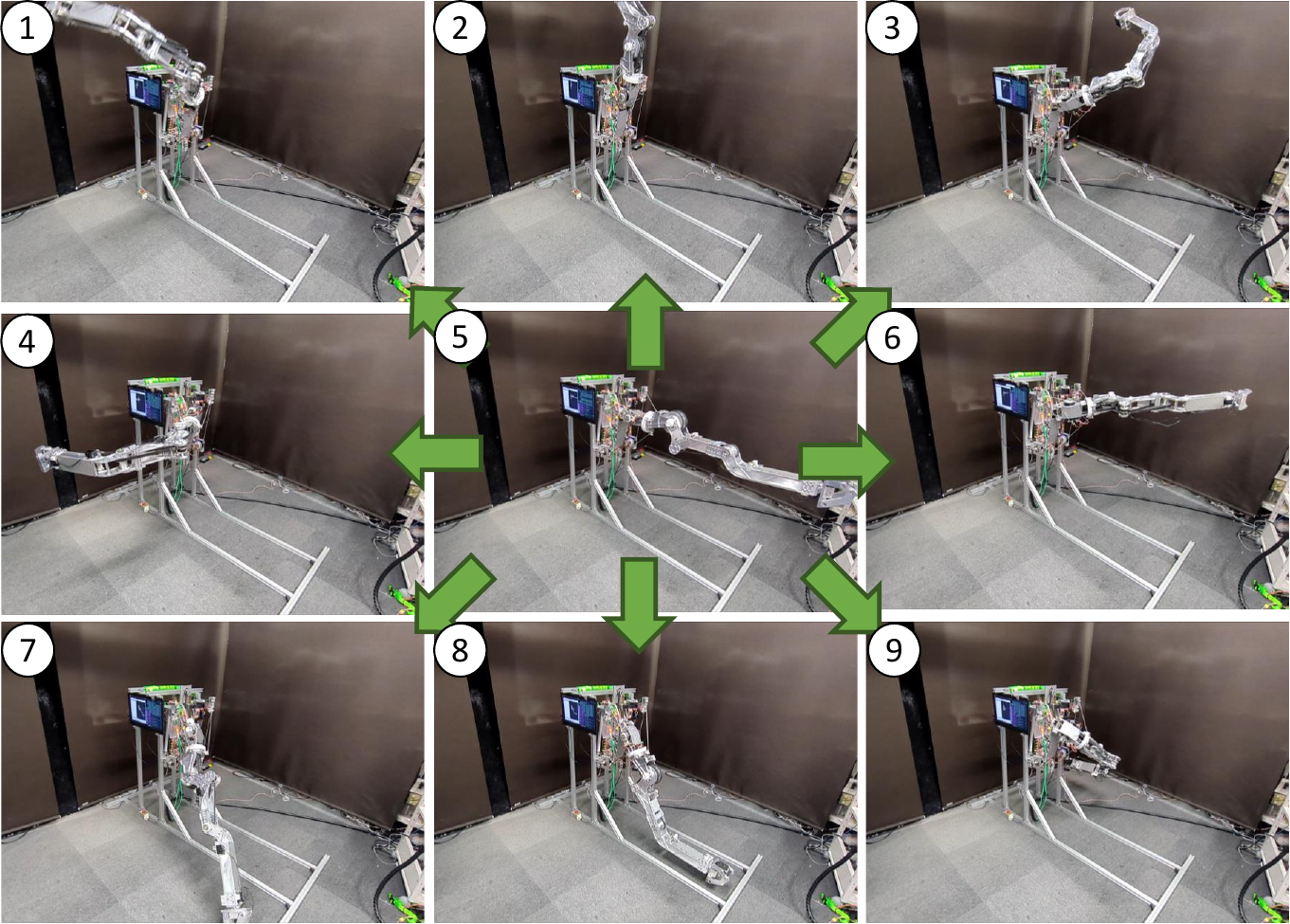}
      \caption{Reachability experiment, moving to various joint angles.}
      \label{figure:Vlimb_data_angle_vector_reachability_9-crop}
\end{figure}

\begin{figure}[tb]
  \centering
      \includegraphics[width=1\columnwidth]{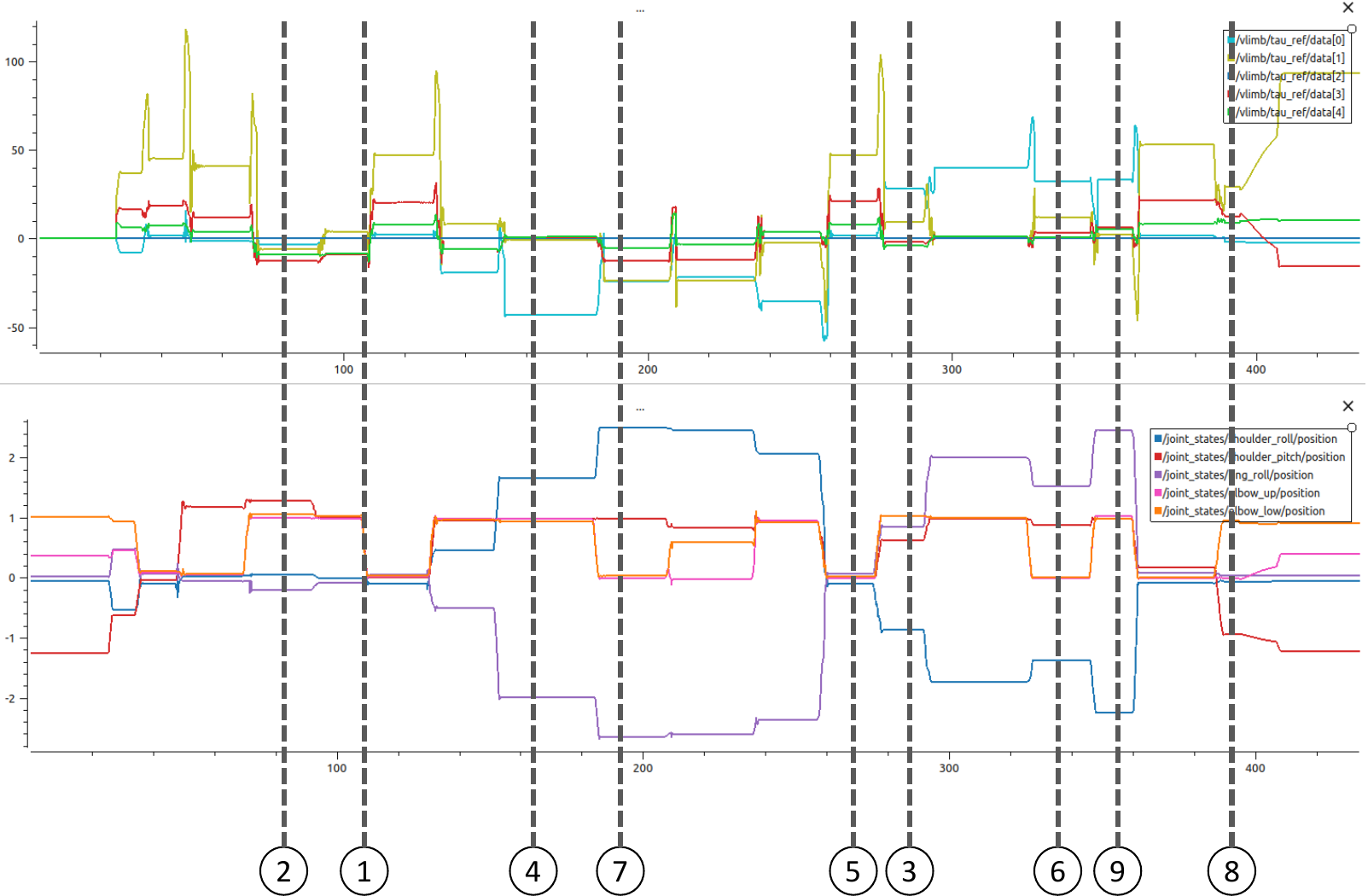}
      \caption{Reachability experiments, the commanded torque at each joint angle.}
      \label{figure:Vlimb_data_angle_vector_reachability_graph-crop}
\end{figure}

\begin{figure*}[tb]
  \centering
      \includegraphics[width=2\columnwidth]{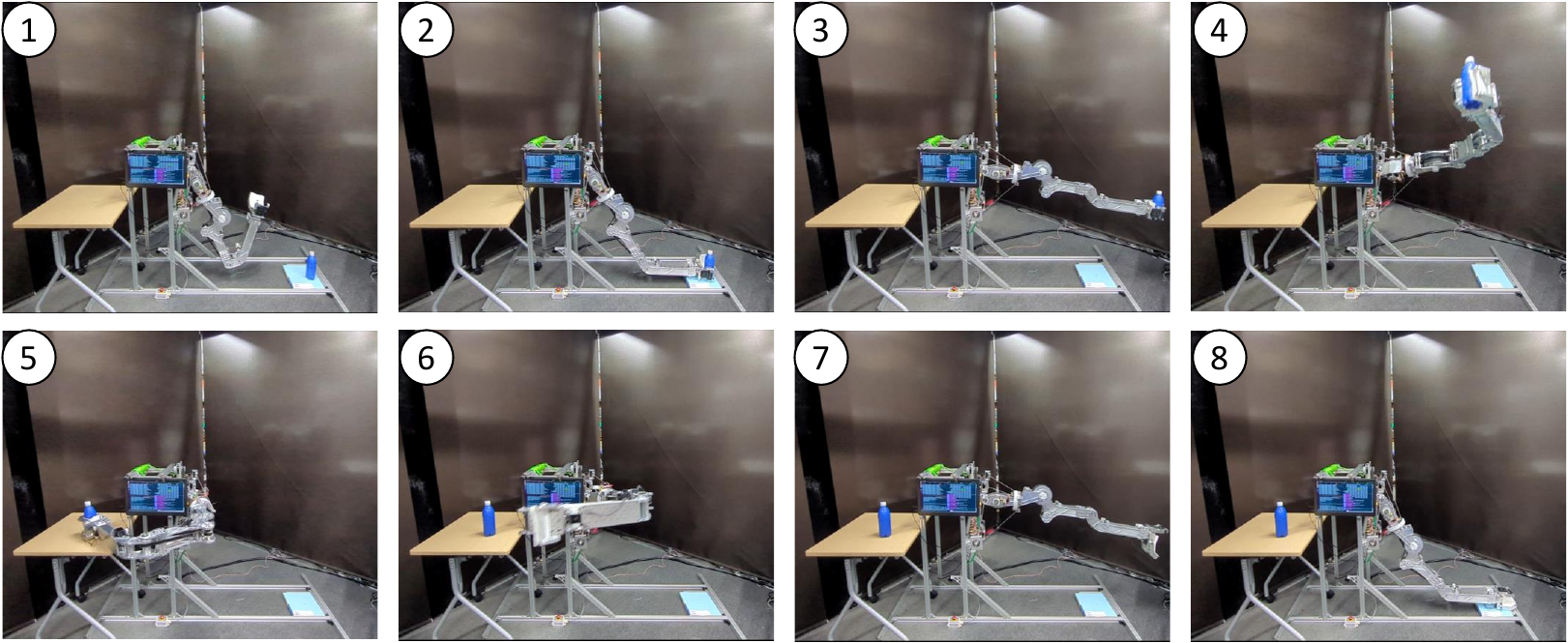}
      \caption{The experiment of Vlimb bringing plastic bottle with water forward from the back posture.}
      \label{figure:experiment_manipulation_8-EN}
\end{figure*}

\begin{figure}[tb]
  \centering
      \includegraphics[width=1\columnwidth]{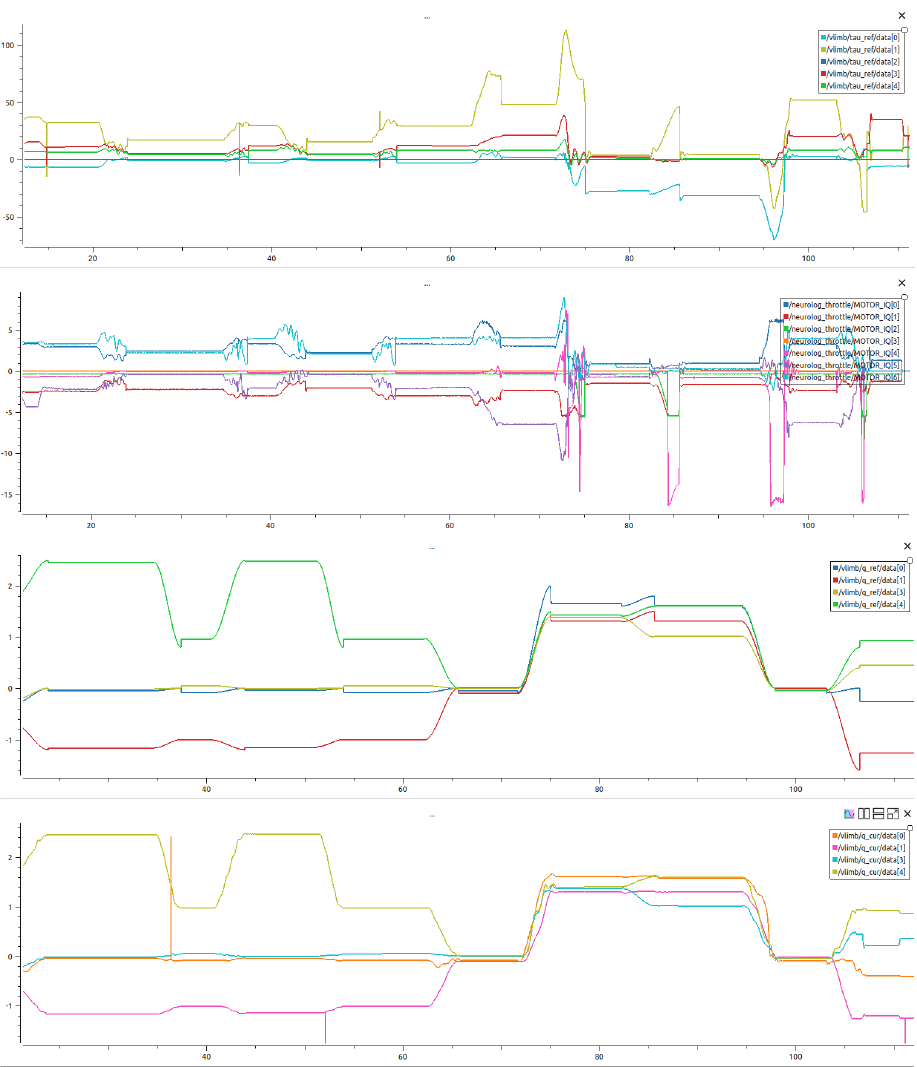}
      \caption{The graphs of the commanded joint angle with the actual joint angle ,and the joint torque commanded with the current value flowing to the motor.}
      \label{figure:experiment_manipulation_graph}
\end{figure}

\begin{figure}[tb]
  \centering
      \includegraphics[width=1\columnwidth]{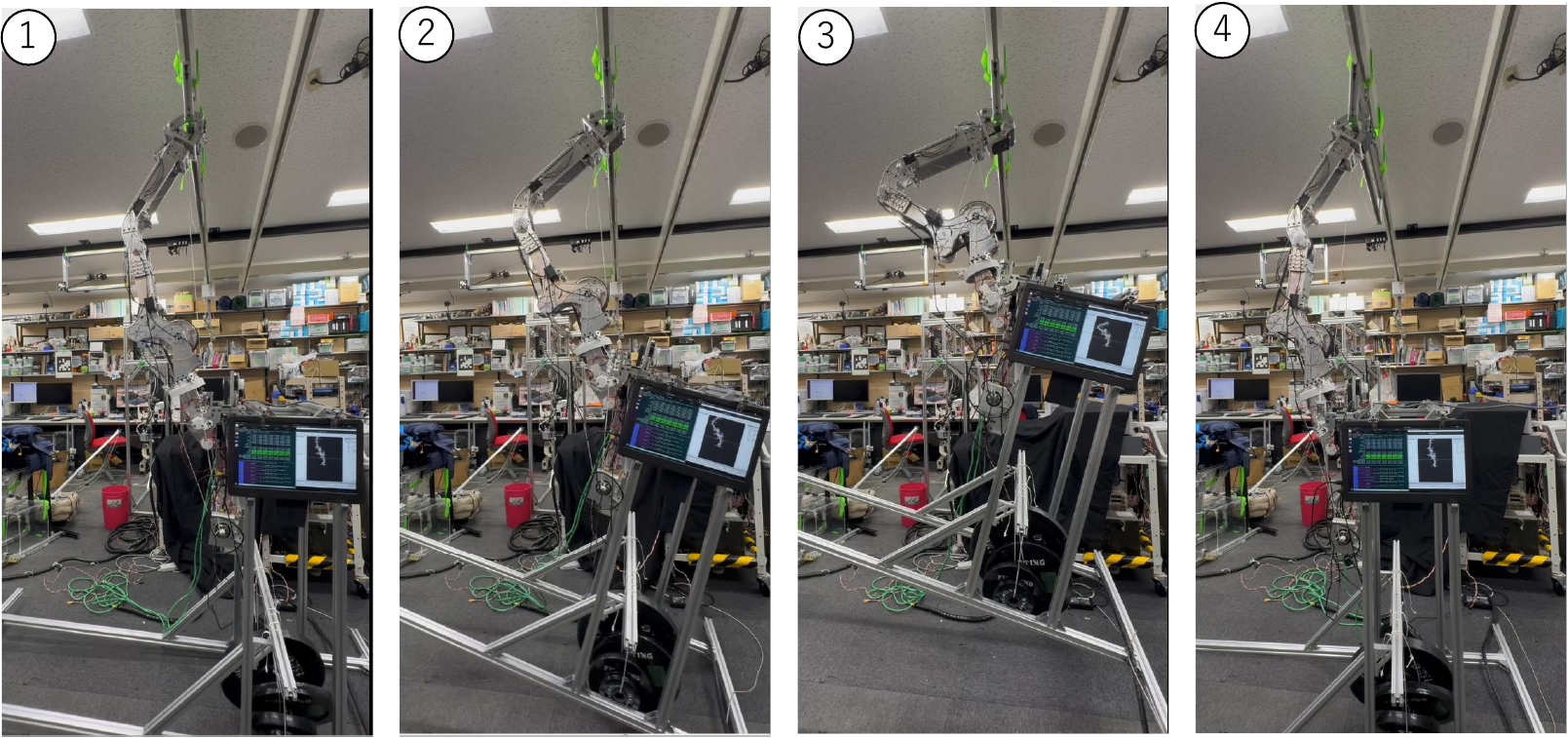}
      \caption{The experiment of Vlimb lifting a 60 kg weight.}
      \label{figure:experiment_lift}
\end{figure}

The experiment confirmed the following two points regarding the created Vlimb:
\begin{itemize}
  \item Commanding joint angle to check reachability in Manipulation Mode: we confirmed whether Vlimb have a wide reachability in Manipulation Mode.
  \item Manipulating light weitght in Manipulation Mode: we confirmed whether Vlimb could manipulate a plastic bottle filled with water (0.5 kg) in Manpulation Mode.
  \item Lifting capacity in Power Mode: we confirmed whether Vlimb could lift a weight equivalent to that of an adult male (60 kg) in Power Mode.
\end{itemize}
For safety reasons, the Vlimb was securely fixed to an aluminum frame during the experiment.
Additionally, power for the motors driving Vlimb and for the control circuits was supplied externally via cables.

\subsection{Reachability Experiment}
In this experiment, we verified whether Vlimb could command appropriate joint torques for each joint.

The behavior when applying various joint angles is illustrated in \figref{figure:Vlimb_data_angle_vector_reachability_9-crop}.
We confirmed that Vlimb could maintain its own weight at various joint angles and posture control is possible at various joint angles. 
Thereby verifying the Vlimb has the wide reachability of the mechanism. 

However, it is worth noting that compensation for friction in each joint was not performed, which might require overcoming the frictional forces to initiate movement.
And from \figref{figure:Vlimb_data_angle_vector_reachability_graph-crop}, it can be confirmed that the commanded torque of the shoulder pitch is overshooting. 
This is considered to be due to latency caused by wire driving.

\subsection{Manipulation Experiment}

In this experiment, a plastic bottle containing water weighing 575 g is manipulated.

\figref{figure:experiment_manipulation_8-EN} illustrates the result of the experiment.
We envision a situation in which the robot grabs an object directly behind a human, brings it in front of the wearer, and passes it to the human.
This shows that the wire-driven wearable robot is capable of a wide range of manipulation movements.

The graphs plotting the joint angles and actual angles, as well as the command torque and the current value to the motor in a series of movements are shown in \figref{figure:experiment_manipulation_graph}.
The graph shows that the robot was able to follow the angles when it held an object.
However, some vibrations were observed that could not be confirmed when no object was held.
This may be due to the fact that the weight of the object held in the hand was not taken into account.

\subsection{Lifting Experiment}
In this experiment, the Vlimb was operated in power mode by reconfiguring the wire routing, rather than in manipulation mode. 
A 60 kg weight was lifted using the Vlimb.
A 30 mm diameter steel bar was mounted on the ceiling, and the Vlimb's hand grasped it beforehand.

The lifting process is illustrated in \figref{figure:experiment_lift}.
The experimental results demonstrated that the Vlimb could lift a weight of 61 kg (77 kg including its own weight) from the ground to approximately 400 mm above it. 
However, it was observed that the Vlimb stopped short of reaching the joint's maximum reachability. 
This phenomenon occurred because bending the joints beyond a certain angle caused friction between the belt and wires, leading to a halt.

}%
{%
  実験では作成したVlimbが以下の三点を満たすことを確認する．
  \begin{itemize}
    \item マニピュレーションモードにおいて,想定しているリーチャビリティの範囲を有すること
    \item マニピュレーションモードにおいて,実際の物体操作が可能であることを確認する
    \item パワーモードにおいて,成人男性相当の重量(60 kg)を持ち上げられること
  \end{itemize}

今回は安全を考慮し，Vlimbをアルミフレームの台に固定して実験を行った．
Vlimbを駆動するモータ用の電源及び制御回路用の電源はどちらも外部からケーブルを用いて供給を行った．

\subsection{リーチャビリティ実験}
本実験ではVlimbのリーチャビリティを確認した.
様々な関節角度へ移動し，姿勢を保てることを示すことによってVlimbの可動域を示す．



各姿勢においての各関節のトルクとその角度をプロットしたグラフを\figref{figure:experiment_reachability}に示す．
様々な関節角度での姿勢制御が可能であることが確認でき,これにより機体の広いリーチャビリティを確認できた．

しかし，各関節にかかっている摩擦について補償を行っていないため，定常誤差が乗り一定の角度に収束するのに時間がかかった.
特に根元のピッチ方向において，グラフよりワイヤによるレイテンシが確認でき，停止制御にてオーバーシュートが起きている.

\begin{figure}[tb]
    \centering
        \includegraphics[width=0.95\columnwidth]{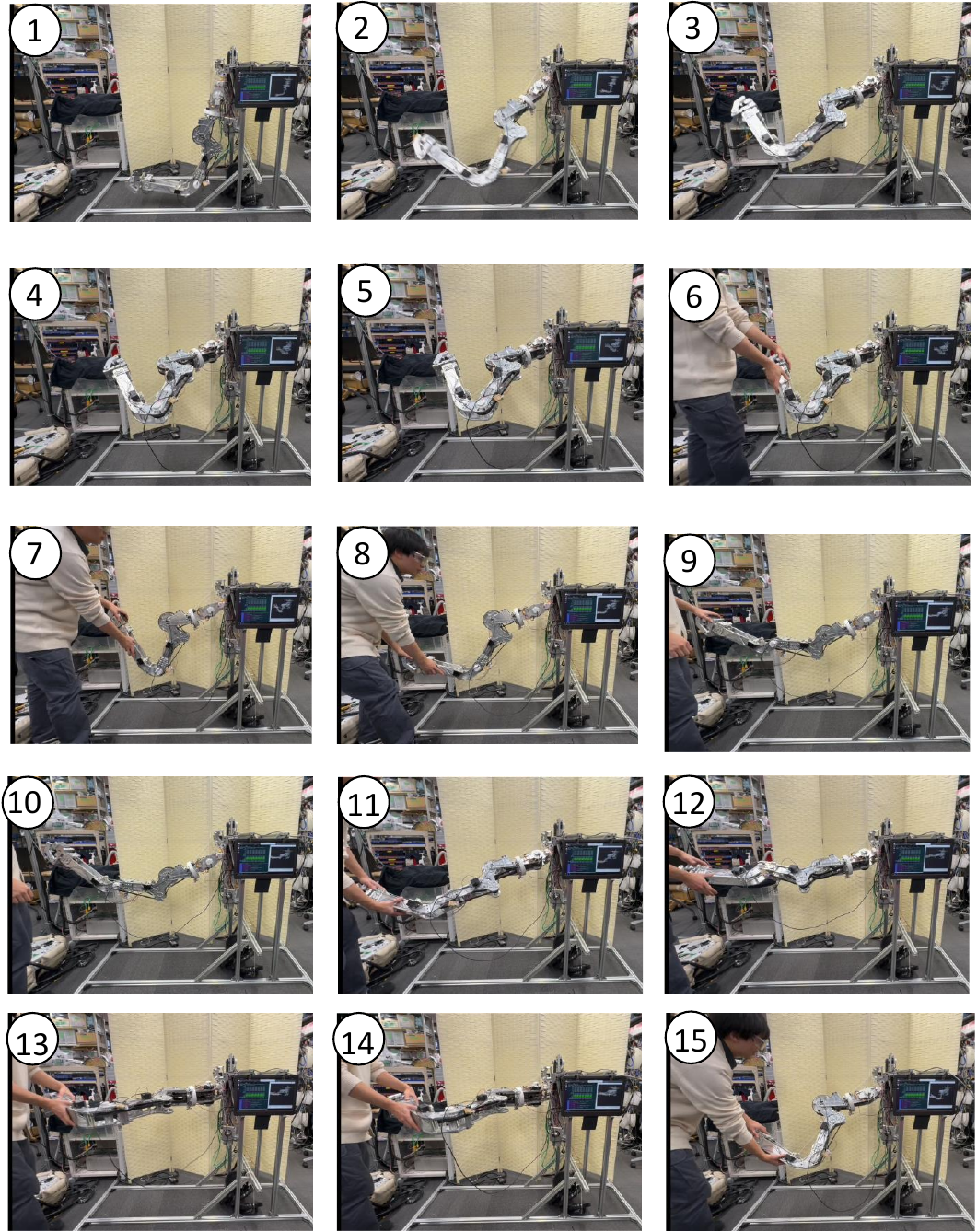}
        \caption{Gravity compensation experiment}
        \label{figure:experiment_gravity_compensation}
\end{figure}

\subsection{物体操作実験}
本実験では，重量575 gの水入りのペットボトルのマニピュレーションを行う.

実際のマニピュレーション動作の実験の様子を図に示す．
人間の真後ろにある物体を掴み、装着している人間の前に持ってきて,人に渡す状況を想定しています．
これにより、ワイヤ駆動ウェアラブルロボットが広い範囲でのマニピュレーション動作が可能であると示せた．

また，一連動作においての関節角度と実際の角度、及び指令トルクとモーターへの電流値をプロットしたグラフをfigに示す．
これにより，ものを持った際にお角度追従ができていることが確認できた．
しかし，一部では物体を持っていないときに確認できなかった振動が生じていた．
これは，手先に持っている物体の重量を考慮に入れていないことに起因すると考えられる．

\begin{figure}[tb]
    \centering
        \includegraphics[width=0.95\columnwidth]{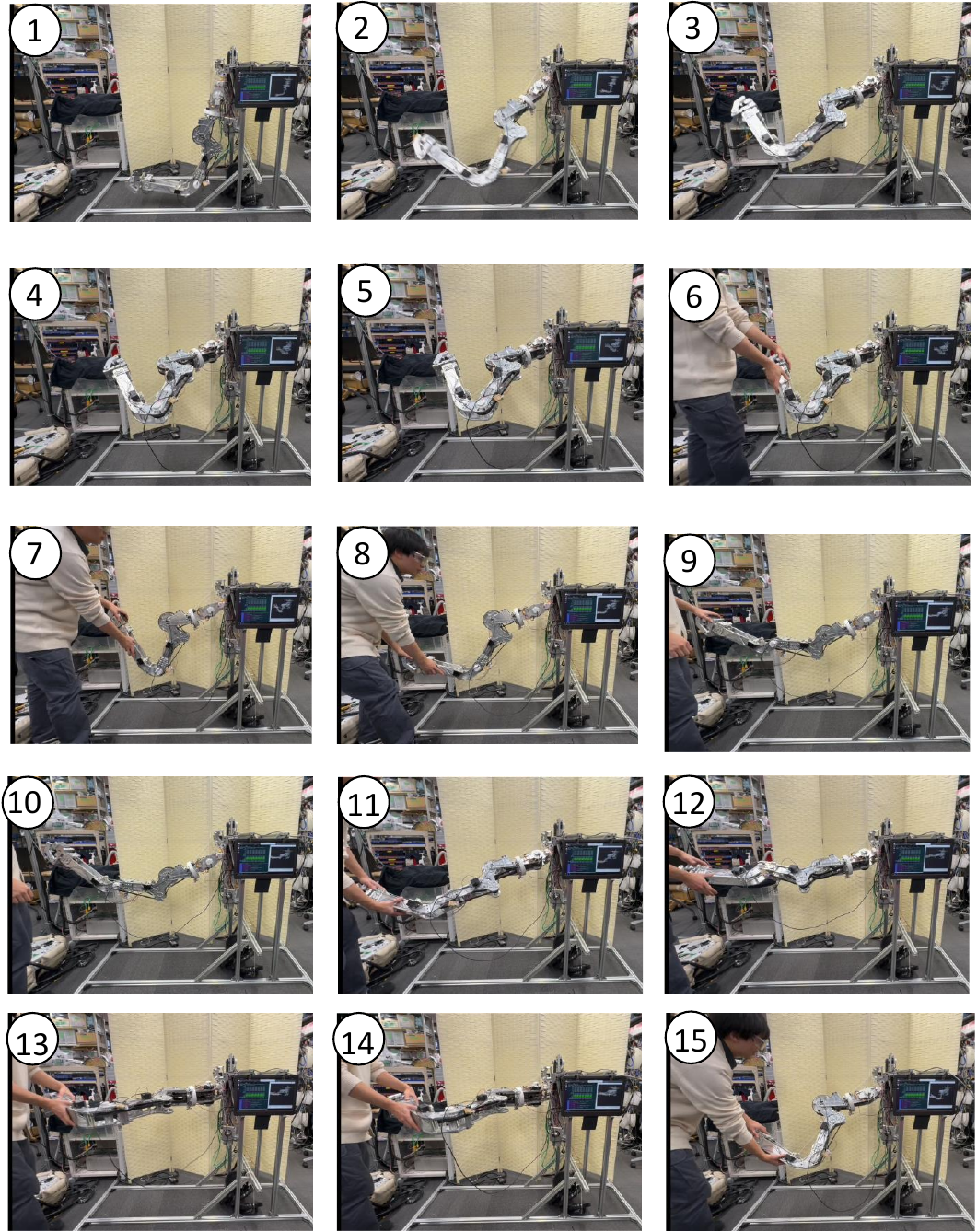}
        \caption{Gravity compensation experiment}
        \label{figure:experiment_manipulation}
\end{figure}

\subsection{持ち上げ実験}
本実験ではVlimbで60 kgを持ち上げた．

天井に直径30 mmの鉄棒を設置し，Vlimbのハンドで事前に掴ませた．
この時ワイヤを経由点から外し，パワーモードで行う．
Vlimb本体重量を除いた61 kgを持ち上げた．

実際に持ち上げた様子を\figref{figure:experiment_lift}に示す．

実験結果より，Vlimbによって重量60 kgの物体を地面から400 mm程度持ち上げ可能であることが確認できた．
一方でVlimbが持ち上がったあと，関節可動限界に到達する前に止まっていることが確認できた．
これは一定以上関節角度を曲げてしまうと，ベルトとワイヤ同士の接触が発生し，その摩擦により静止してしまうのが原因と考えられる．

\begin{figure}[tb]
    \centering
        \includegraphics[width=1\columnwidth]{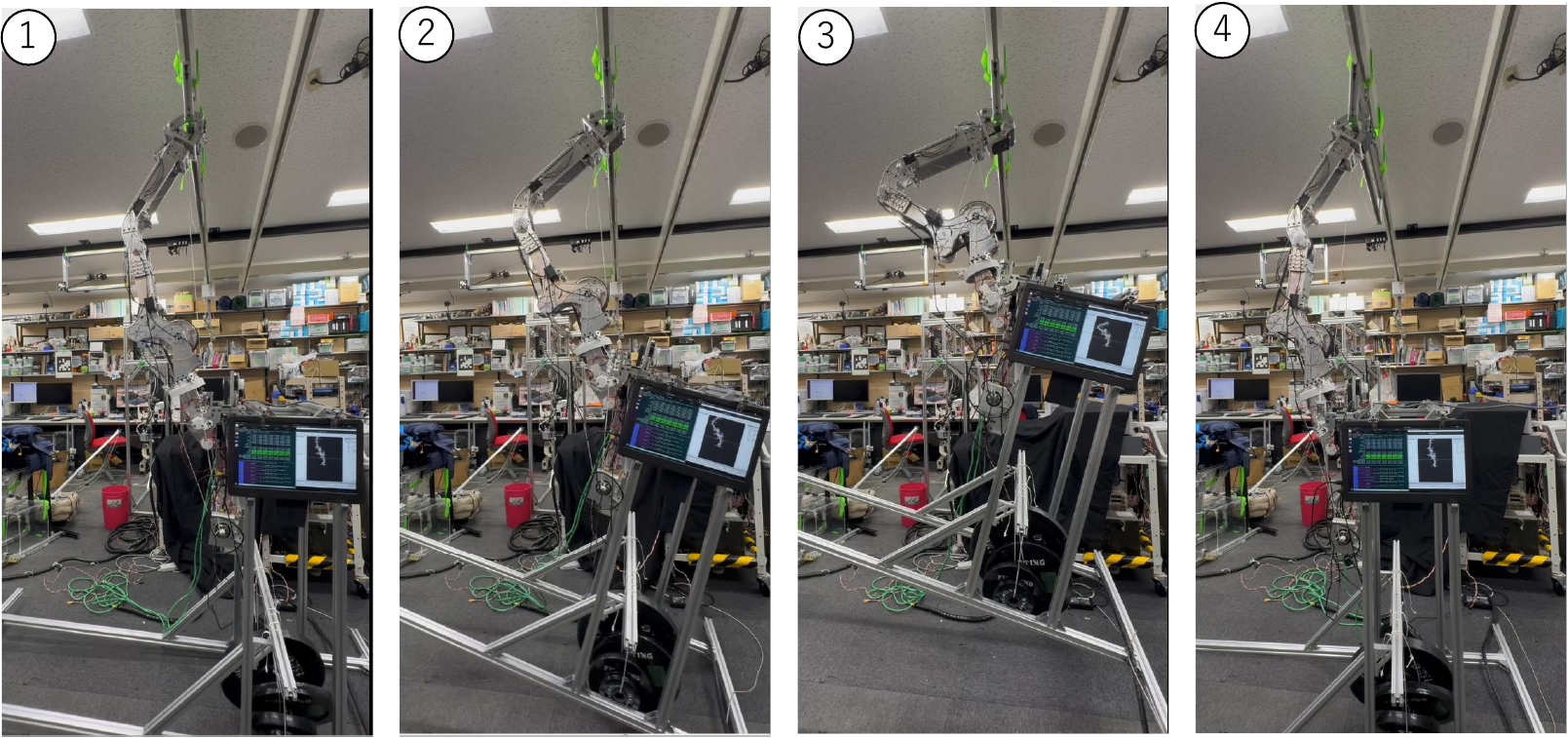}
        \caption{Weight lifting experiment 60 kg}
        \label{figure:experiment_lift}
\end{figure}
}%

\section{Conclusion} \label{sec:conclusion}
\switchlanguage%
{%
In this study, we proposed Vlimb, a bodily extension wearable robot capable of surpassing the powerfulness and reachability of human limbs. 
Vlimb features switchable wire waypoints and a passive ring mechanism. 
By altering the wire waypoints, Vlimb can switch between a Power Mode, capable of exerting enough force to lift humans, and a Manipulation Mode, offering reachability beyond human limbs.
The passive ring mechanism addresses the limitations of wire-driven systems, enabling Vlimb to achieve a wide reachability.
In our experiments, we demonstrated the feasibility of manipulation within a wide reachability by applying gravity compensation torque and specifying appropriate torque for each joint. 
In addition, a manipulation task in which an object is picked up from behind and brought to the front of the human was performed, and it was shown that manipulation is possible.
Furthermore, we showcased Vlimb's powerfulness by lifting a weight equivalent to that of an adult male, 60 kg.

Looking ahead, our future work involves implementing control strategies with safety considerations. 
Subsequently, we aim to attach Vlimb to human subjects and achieve powerful three-dimensional movements, such as parkour, through cooperation with humans. 
}%
{%
本研究では四肢を超える力強さと広い動作範囲を有する身体拡張ウェアラブルロボットVlimbを提案した．
Vlimbは切り替え可能なワイヤ経由点と, 受動リング機構を有する.
ワイヤ経由点を変更することにより，人が持ち上げられるほどの力を発揮できるパワーモードと人間の四肢以上のリーチャビリティを有するマニピュレーションモードの切り替えが可能となる．
受動リング機構により，ワイヤ駆動において欠点となる限られた動作範囲を解消し，広い動作範囲を実現した．

実験として，広いリーチャビリティを持っていることを確認するため、様々な角度へ関節指令を与え動作範囲を示した．
また,背後にある物体を取りあげ人間の前面に持ってくる状況を想定したマニピュレーションタスクを行い,マニピュレーションが可能であることを示した.
さらには,成人男性相当の重量である60 kgの持ち上げ,力強さを示した．

今後の展望として安全性を考慮した制御を実装したのち，実際に人に取り付け，人との協調によるパルクールなどの力強い三次元動作の実現を目指す．

\footnotesize
}%

{
  \bibliographystyle{IEEEtran}
  \bibliography{main}

\begin{thebibliography}{10}
\providecommand{\url}[1]{#1}
\csname url@rmstyle\endcsname
\providecommand{\newblock}{\relax}
\providecommand{\bibinfo}[2]{#2}
\providecommand\BIBentrySTDinterwordspacing{\spaceskip=0pt\relax}
\providecommand\BIBentryALTinterwordstretchfactor{4}
\providecommand\BIBentryALTinterwordspacing{\spaceskip=\fontdimen2\font plus
\BIBentryALTinterwordstretchfactor\fontdimen3\font minus
  \fontdimen4\font\relax}
\providecommand\BIBforeignlanguage[2]{{%
\expandafter\ifx\csname l@#1\endcsname\relax
\typeout{** WARNING: IEEEtran.bst: No hyphenation pattern has been}%
\typeout{** loaded for the language `#1'. Using the pattern for}%
\typeout{** the default language instead.}%
\else
\language=\csname l@#1\endcsname
\fi
#2}}

\bibitem{Hayashi2015HAL}
T.~Hayashi, H.~Kawamoto, and Y.~Sankai, ``Control method of robot suit hal
  working as operator's muscle using biological and dynamical information,'' in
  \emph{2005 IEEE/RSJ International Conference on Intelligent Robots and
  Systems}, 2005, pp. 3063--3068.

\bibitem{CYBERDYNEHAL}
``{CYBERDYNE HAL},''
  \url{https://www.cyberdyne.jp/english/products/HAL/index.html}.

\bibitem{MusclesSuits}
``{INNOPHYS MusclesSuits Every},'' \url{https://innophys.net/musclesuit/}.

\bibitem{Yamamura2023JizaiArm}
N.~Yamamura, D.~Uriu, M.~Muramatsu, Y.~Kamiyama, Z.~Kashino, S.~Sakamoto,
  N.~Tanaka, T.~Tanigawa, A.~Onishi, S.~Yoshida, \emph{et~al.}, ``{Social
  Digital Cyborgs: The Collaborative Design Process of JIZAI ARMS},'' in
  \emph{Proceedings of the 2023 CHI Conference on Human Factors in Computing
  Systems}, 2023, pp. 1--19.

\bibitem{yang2021supernumerary}
B.~Yang, J.~Huang, X.~Chen, C.~Xiong, and Y.~Hasegawa, ``{Supernumerary robotic
  limbs: a review and future outlook},'' \emph{IEEE Transactions on Medical
  Robotics and Bionics}, vol.~3, no.~3, pp. 623--639, 2021.

\bibitem{parietti2014supernumerary}
F.~Parietti and H.~H. Asada, ``{Supernumerary robotic limbs for aircraft
  fuselage assembly: body stabilization and guidance by bracing},'' in
  \emph{Proceedings of the 2014 IEEE International Conference on Robotics and
  Automation}.\hskip 1em plus 0.5em minus 0.4em\relax IEEE, 2014, pp.
  1176--1183.

\bibitem{parietti2016supernumerary}
F.~Parietti and H.~Asada, ``Supernumerary robotic limbs for human body
  support,'' \emph{IEEE Transactions on Robotics}, vol.~32, no.~2, pp.
  301--311, 2016.

\bibitem{sasaki2017metalimbs}
T.~Sasaki, M.~Y. Saraiji, C.~L. Fernando, K.~Minamizawa, and M.~Inami,
  ``{MetaLimbs: multiple arms interaction metamorphism},'' in \emph{ACM
  SIGGRAPH 2017 Emerging Technologies}, 2017, pp. 1--2.

\bibitem{amano2019development}
K.~Amano, Y.~Iwasaki, K.~Nakabayashi, and H.~Iwata, ``{Development of a
  three-fingered jamming gripper for corresponding to the position error and
  shape difference},'' in \emph{Proceedings of the 2019 IEEE International
  Conference on Soft Robotics}, 2019, pp. 137--142.

\bibitem{iwasaki2022ExtraLimb}
Y.~Iwasaki, S.~Takahashi, and H.~Iwata, ``Comparison of operating method of
  extra limbs in dual tasks: point and path instruction,'' in \emph{2022
  IEEE/SICE International Symposium on System Integration (SII)}.\hskip 1em
  plus 0.5em minus 0.4em\relax IEEE, 2022, pp. 171--176.

\bibitem{Veronneau20203-DOF}
C.~V{\'e}ronneau, J.~Denis, L.-P. Lebel, M.~Denninger, V.~Blanchard, A.~Girard,
  and J.-S. Plante, ``{Multifunctional Remotely Actuated 3-DOF Supernumerary
  Robotic Arm Based on Magnetorheological Clutches and Hydrostatic Transmission
  Lines},'' \emph{IEEE Robotics and Automation Letters}, vol.~5, no.~2, pp.
  2546--2553, 2020.

\bibitem{kawamura1997development}
S.~Kawamura, W.~Choe, S.~Tanaka, and H.~Kino, ``{Development of an ultrahigh
  speed robot FALCON using parallel wire drive systems},'' \emph{Journal of the
  Robotics Society of Japan}, vol.~15, no.~1, pp. 82--89, 1997.

\bibitem{Friesen2018ICRA}
J.~M. Friesen, J.~L. Dean, T.~Bewley, and V.~Sunspiral, ``{A
  Tensegrity-Inspired Compliant 3-DOF Compliant Joint},'' in \emph{Proceedings
  of the 2018 IEEE International Conference on Robotics and Automation}, 2018,
  pp. 3301--3306.

\bibitem{Temma}
T.~Suzuki, M.~Bando, K.~Kawaharazuka, K.~Okada, and M.~Inaba, ``Saqiel:
  Ultra-light and safe manipulator with passive 3d wire alignment mechanism,''
  \emph{IEEE Robotics and Automation Letters}, vol.~9, no.~4, pp. 3720--3727,
  2024.

\bibitem{carpentier2019pinocchio}
J.~Carpentier, G.~Saurel, G.~Buondonno, J.~Mirabel, F.~Lamiraux, O.~Stasse, and
  N.~Mansard, ``{The Pinocchio C++ library -- A fast and flexible
  implementation of rigid body dynamics algorithms and their analytical
  derivatives},'' in \emph{Proceedings of the IEEE International Symposium on
  System Integrations}, 2019.

\bibitem{Kengoro2016Jacobian}
M.~Kawamura, S.~Ookubo, Y.~Asano, T.~Kozuki, K.~Okada, and M.~Inaba, ``{A
  joint-space controller based on redundant muscle tension for multiple DOF
  joints in musculoskeletal humanoids},'' in \emph{Proceedings of the 2016
  IEEE-RAS 16th International Conference on Humanoid Robots}, 2016, pp.
  814--819.

\end{thebibliography}
}

\end{document}